\documentclass[sigconf,nonacm]{acmart}

\usepackage{caption}
\usepackage{subcaption}

\AtBeginDocument{%
  \providecommand\BibTeX{{%
    \normalfont B\kern-0.5em{\scshape i\kern-0.25em b}\kern-0.8em\TeX}}}

\begin{document}
\newcommand{\myx}{{\boldsymbol{x}}}
\newcommand{\myc}{{\boldsymbol{c}}}
\newcommand{\myC}{{\boldsymbol{C}}}
\newcommand{\myn}{{\boldsymbol{n}}}
\newcommand{\myo}{{\boldsymbol{o}}}
\newcommand{\myv}{{\boldsymbol{v}}}
\newcommand{\myp}{{\boldsymbol{p}}}

\title{VoxNeuS: Enhancing Voxel-Based Neural Surface Reconstruction via Gradient Interpolation}


\author{Sidun Liu}
\orcid{0000-0001-7715-4698}
\affiliation{%
 \institution{National University of Defence Technology}
 \city{Changsha}
 \country{China}}
\email{liusidun@nudt.edu.cn}
\author{Peng Qiao}
\orcid{0000-0001-6752-7892}
\affiliation{%
 \institution{National University of Defence Technology}
 \city{Changsha}
 \country{China}
}
\email{pengqiao@nudt.edu.cn}
\author{Zongxin Ye}
\affiliation{%
 \institution{National University of Defence Technology}
 \city{Changsha}
 \country{China}
}
\author{Wenyu Li}
\affiliation{%
 \institution{National University of Defence Technology}
 \city{Changsha}
 \country{China}
}
\author{Yong Dou}
\orcid{0000-0002-1256-8934}
\affiliation{%
 \institution{National University of Defence Technology}
 \city{Changsha}
 \country{China}
}
\email{yongdou@nudt.edu.cn}


\begin{abstract}
  Neural Surface Reconstruction learns a Signed Distance Field~(SDF) to reconstruct the 3D model from multi-view images. Previous works adopt voxel-based explicit representation to improve efficiency. However, they ignored the gradient instability of interpolation in the voxel grid, leading to degradation on convergence and smoothness. Besides, previous works entangled the optimization of geometry and radiance, which leads to the deformation of geometry to explain radiance, causing artifacts when reconstructing textured planes.
  In this work, we reveal that the instability of gradient comes from its discontinuity during trilinear interpolation, and propose to use the interpolated gradient instead of the original analytical gradient to eliminate the discontinuity. Based on gradient interpolation, we propose VoxNeuS, a lightweight surface reconstruction method for computational and memory efficient neural surface reconstruction. Thanks to the explicit representation, the gradient of regularization terms, i.e. Eikonal and curvature loss, are directly solved, avoiding computation and memory-access overhead.
  Further, VoxNeuS adopts a geometry-radiance disentangled architecture to handle the geometry deformation from radiance optimization.
  The experimental results show that VoxNeuS achieves better reconstruction quality than previous works. The entire training process takes 15 minutes and less than 3 GB of memory on a single 2080ti GPU.

\end{abstract}





\maketitle

\section{Introduction}

\begin{figure}[t]
    \centering
    \includegraphics[width=.8\linewidth]{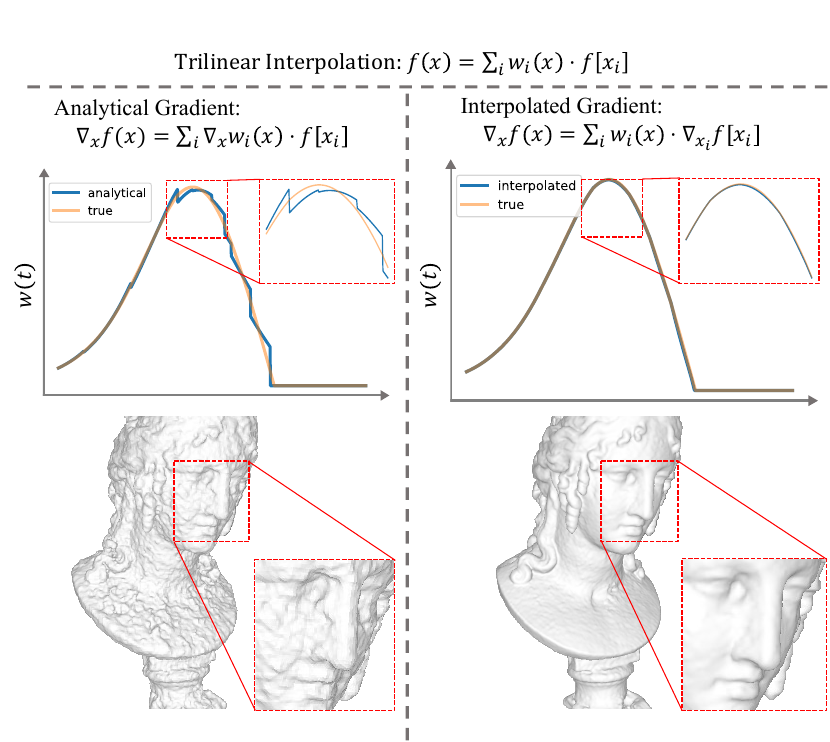}
    \caption{Previous grid-based methods adopt trilinear interpolation to retrieve SDF values and analytically solve SDF gradient. However, the gradient discontinuity leads to glitches in volume rendering weights, thus degrading the reconstruction quality. Our proposed interpolated gradient eliminates the discontinuity and smooths the reconstructed surface without additional computation overhead.}
    \label{fig:teaser}
    \vspace{-5mm}
\end{figure}


3D surface reconstruction aims to recover dense geometric scene structures from multiple images observed at different viewpoints~\cite{hartley2003multiple}. Recently, the success of Neural Radiance Field (NeRF)~\cite{mildenhall2021nerf} on the novel view synthesis task proves the potential of differentiable volume rendering on fitting a density field.
NeuS~\cite{wang2021neus}, as a representative work, uses a neural network to encode the SDF.
However, the implicit representation of SDF leads to a long training time, i.e. about 8 hours for a static object.

As proved in NeRF, explicit volumetric representation achieves higher efficiency and quality than implicit representation~\cite{muller2022instant,sun2022direct,fridovich2022plenoxels}. However, in surface reconstruction tasks, straightforward implementations result in poor geometry and undesirable noise due to the high-frequency nature of explicit volumetric representations and the multi-solution nature of volume rendering. Additional regularization is added for a reasonable surface. Voxurf~\cite{wu2022voxurf}, which directly stores SDF values in a dense grid~\cite{sun2022direct}, designs a dual color network to implicitly regularize the geometry, and uses total variation loss for surface smoothness. Instant NSR\cite{zhao2022human} and NeuS2~\cite{wang2023neus2}, which take iNGP's multi-resolution hash grid~\cite{muller2022instant} as the backbone, utilize multi-resolution architecture for progressive training. 

These works all use grids to replace neural networks, but ignore the impact of gradients on NeuS algorithm. In volumetric representation, the SDF value or latent feature is from the trilinear interpolation of neighbouring vertices. However, the trilinear interpolation is \textit{zero order continuous} only,
which means that its gradient is not continuous at the grid boundary. This problem hinders convergence and the smoothness of surface, as shown in Fig.~\ref{fig:teaser}.


\begin{figure*}[t]
\centering
    \begin{subfigure}[b]{0.24\linewidth}
        \centering
        \includegraphics[width=0.98\linewidth]{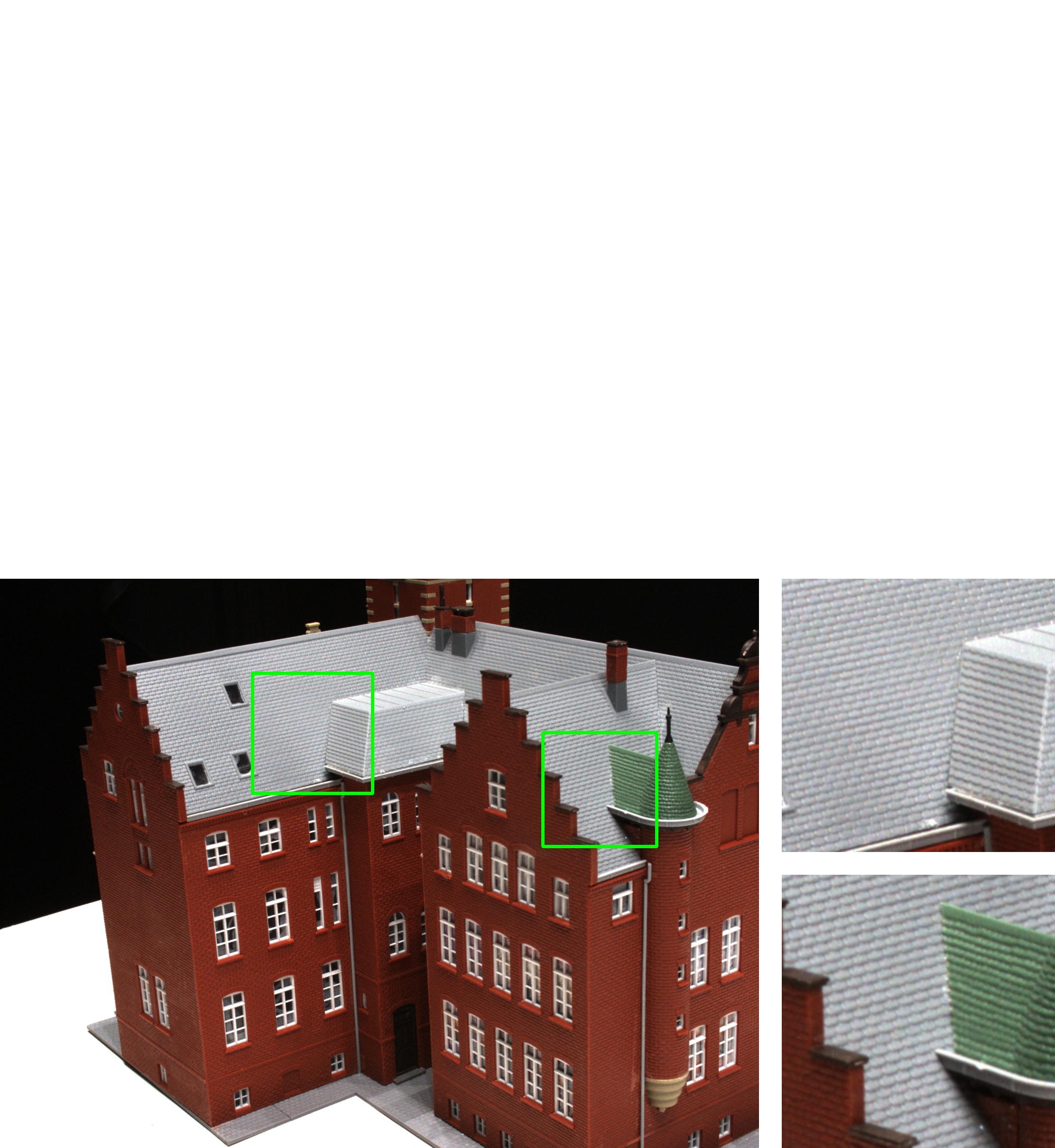}
        \caption{Reference}
    \end{subfigure}
    \begin{subfigure}[b]{0.24\linewidth}
        \centering
        \includegraphics[width=0.98\linewidth]{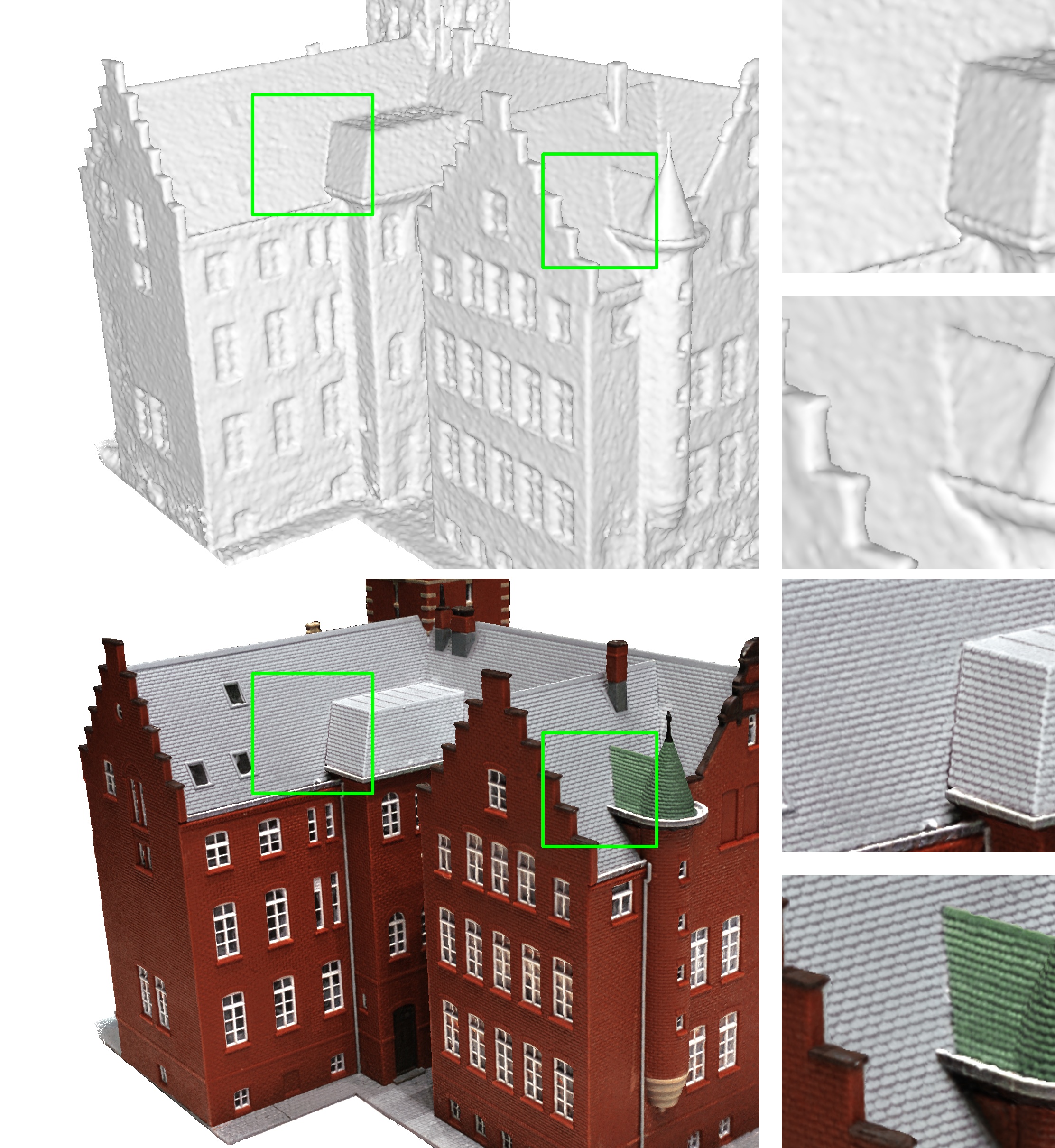}
        \caption{Ours (15 mins; 3GB)}
    \end{subfigure}
    \begin{subfigure}[b]{0.24\linewidth}
        \centering
        \includegraphics[width=0.98\linewidth]{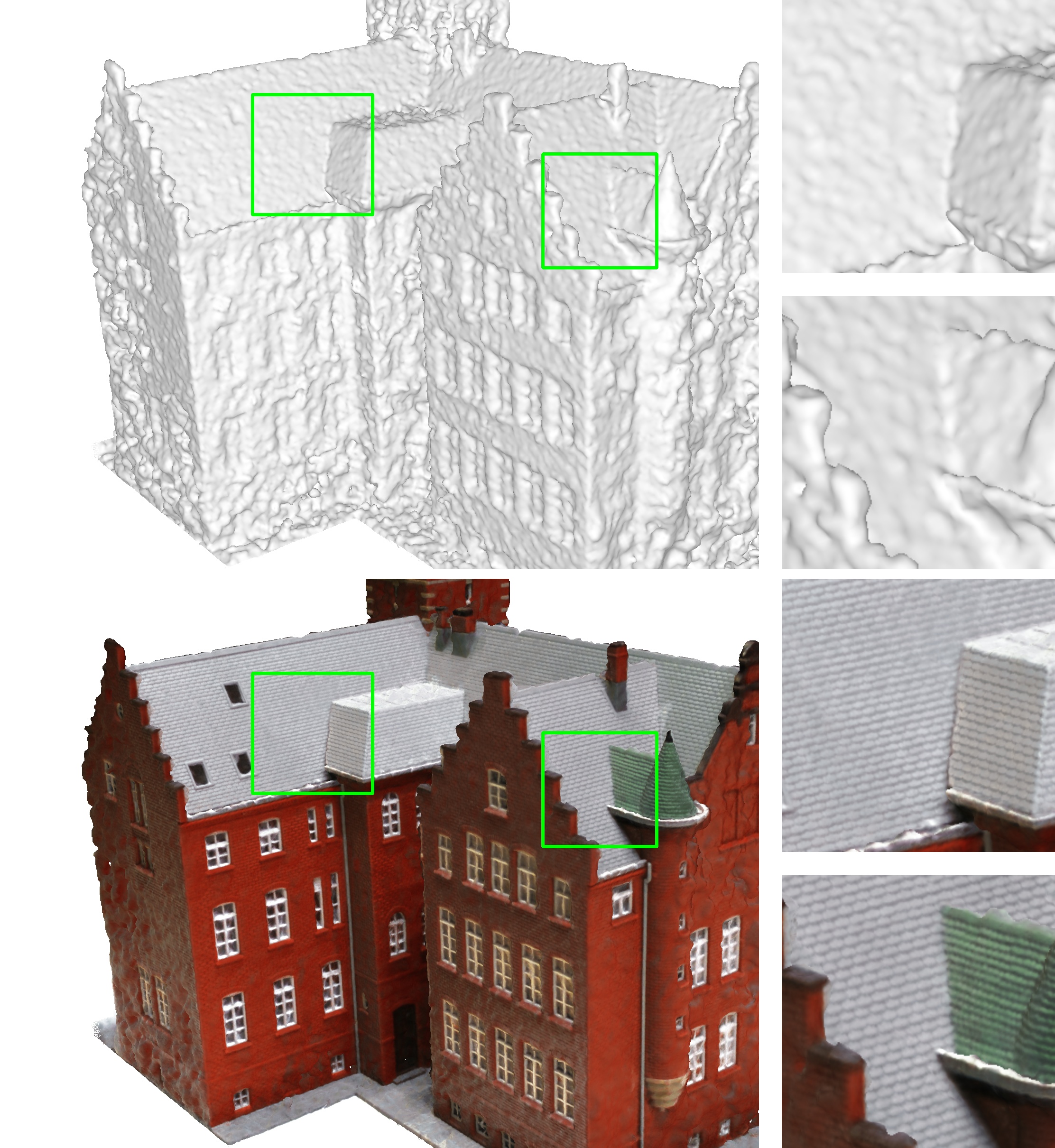}
        \caption{w/o Interp. Grad.}
    \end{subfigure}
    \begin{subfigure}[b]{0.24\linewidth}
        \centering
        \includegraphics[width=0.98\linewidth]{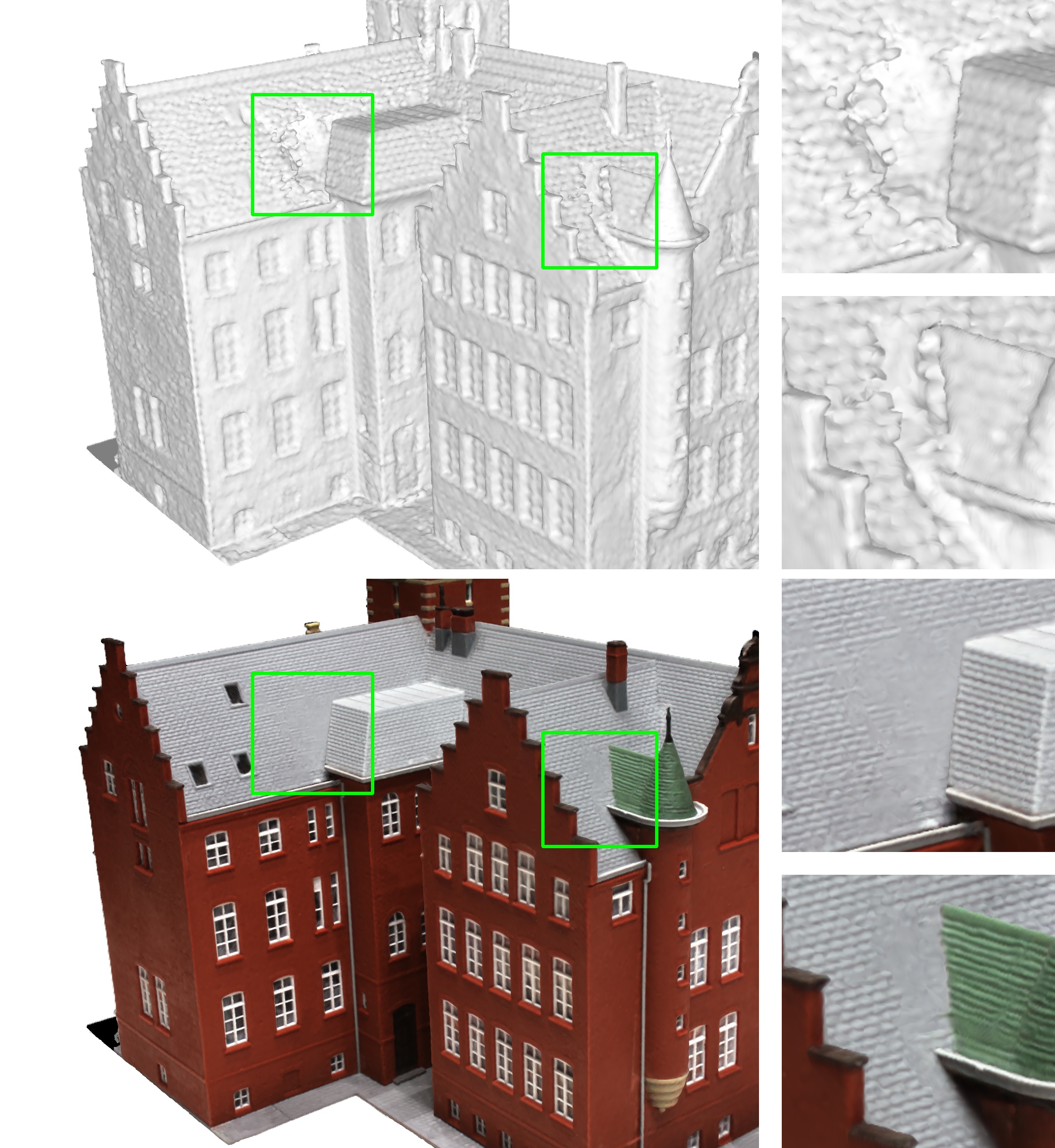}
        \caption{Voxurf (50 mins; 14 GB)}
    \end{subfigure}


    \caption{Reconstruction results from different methods. The running time and GPU memory consumption are tested on a single RTX 3090 for fair comparison, as 2080ti can't meet the memory requirement of Voxurf\cite{wu2022voxurf}. (b) Our proposed VoxNeuS replaces analytical gradient with interpolated gradient to handle the gradient discontinuity in trilinear interpolation. Combined with a simple network structure, our method achieves efficient and high-quality reconstruction. (c) When the interpolation gradient is removed and other structures remain unchanged, the reconstruction quality drops drastically. (d) Voxurf adopts dual color network and other tricks to ease the optimization of the SDF grid, which leads to some artifacts on the surface.}
\label{fig:comp1}
\end{figure*}

When SDF is encoded by a hash grid or other hybrid representation, we need to use numerical gradients~\cite{li2023neuralangelo} to eliminate gradient discontinuity, bringing the multiplied amount of the computation and memory. In contrast, when the SDF is explicitly stored in the vertices of a dense grid~\cite{wu2022voxurf}, we can directly interpolate the gradients of the vertices to ensure gradient continuity.

Based on the above analysis, we propose VoxNeuS, an efficient and lightweight surface reconstruction method that directly optimizes the SDF dense grid. In VoxNeuS, the analytical gradient is replaced with interpolated gradient to ensure continuity without additional computational overhead. With gradient interpolation, the gradient of vertices is first estimated with central finite differences. Then the target gradient is interpolated with these estimated vertex gradients to ensure zero order continuity of gradient. In addition, more designs to improve reconstruction efficiency and quality include: 1) We adopt a geometry-radiance disentangled network architecture to eliminate the impact of surface texture on geometry when fitting radiance; 2) We directly apply the gradient of the SDF regularization, i.e. Eikonal loss~\cite{gropp2020implicit} and curvature loss~\cite{li2023neuralangelo}, on vertices to avoid the overhead of forward and backward pass; 3) We leverage a progressive super-resolution of SDF grid to balance the coarse shape and fine details; 4) Critical operators like interpolation and vertices regularization are implemented in CUDA to accelerate computation and save memory, and the sparsity of gradients is considered to improve efficiency further.

The experiments are conducted on the DTU~\cite{jensen2014large} and BlendedMVS~\cite{yao2020blendedmvs} datasets for quantitative and qualitative evaluations. The results demonstrate that VoxNeuS achieves a lower Chamfer Distance on the DTU benchmark than competitive volumetric-based methods, i.e. Voxurf and NeuS2. Our contributions are as following:




\begin{itemize}
    \item We reveal that the key reason that hinders the convergence and smoothness of the NeuS in volumetric representation is the gradient discontinuity caused by trilinear interpolation.
    \item Our method replaces the analytical gradient of SDF with an interpolated gradient to handle the gradient discontinuity without any computational overhead.
    \item Our method converges well without an auxiliary foreground mask and achieves better quantitative results than competitive methods while the computational and memory efficiency improves. The training finishes in 15 minutes on a single 2080ti GPU with less than 3GB of memory cost.
\end{itemize}

\section{Related Work}

\noindent\textbf{Neural Surface Reconstruction.}
Recently popular neural surface reconstruction methods use Signed Distance Function~(SDF)~\cite{green2007improved} to represent the surface. Early surface rendering methods like DVR~\cite{niemeyer2020differentiable} and IDR~\cite{yariv2020multiview} seek the ray intersection point with the surface and optimize its color and coordinate. Inspired by the volume rendering, recent works densely sample points along the ray and blend their colors as ray color. NeuS~\cite{wang2021neus} and VolSDF~\cite{yariv2021volume} formulate equivalent density/opacity in volume rendering with SDF values. The following works~\cite{darmon2022improving,wang2022hf,zhuang2023anti,zhang2023towards,tian2023superudf,ye2023nef} make efforts to improve the quality and efficiency. Notably, a representative method NeuS introduces the derivative of the SDF value to the coordinate as the normal in the forward pass. This mechanism benefits the convergence when the SDF is represented by an MLP. However, in the context of grid-based explicit representation where trilinear interpolation is adopted for SDF value prediction, the inaccurate estimation of gradient and the discontinuity at junction surfaces hinder the reconstruction quality drastically.
Another line of work adopts recently popular 3DGS~\cite{kerbl20233d} for point-based rendering and surface reconstruction~\cite{guedon2023sugar,huang20242d}. However, their efficiency and precision are still unsatisfactory compared to SDF-based rendering.

\noindent\textbf{Volumetric Representation for Surface Reconstruction.}
The Signed Distance Function (SDF) was originally represented with a dense grid when proposed~\cite{green2007improved}, and neural rendering methods replace it with an MLP for easier optimization, but it leads to low efficiency during reconstruction and rendering. Therefore, volumetric-based hybrid representation is adopted for better efficiency and model capacity. Vox-surf~\cite{li2022vox} uses pruned feature grid to encode SDF and color and decodes them with a shallow MLP. NeuDA~\cite{cai2023neuda} maintains the hierarchical anchor grids where each vertex stores a 3D position instead of the direct feature. An MLP is still required to map the deformed coordinates to geometry. The multi-resolution hash grid, proposed by iNGP~\cite{muller2022instant}, is applied to MonoSDF~\cite{yu2022monosdf}, Instant NSR~\cite{zhao2022human}, NeuS2~\cite{wang2023neus2}, etc. for efficient surface reconstruction. PermutoSDF~\cite{rosu2023permutosdf} follows iNGP's multi-resolution structure and designs the permutohedral lattices as a backbone network. PET-NeuS~\cite{wang2023pet} adopts tri-plane representation, along with the self-attention convolution to assist optimization. Unlike the above work, Voxurf~\cite{wu2022voxurf} replaces \textit{grid+MLP} hybrid geometry representation with an explicit dense grid to store SDF values. However, the optimization of explicit SDF grids is more difficult, and Voxurf uses many tricks, such as Gaussian smoothing, dual color networks, etc., to stabilize the optimization. Notably, the geometry feature branch of Voxurf's dual color network predicts the point color based on local SDF values, which causes the geometry to be optimized through the color prediction pass, which violates the NeRF volume rendering convention~\cite{mildenhall2021nerf} and results in undesirable surface.

The above method extends NeuS to grids and uses trilinear interpolation to query features or SDF values. However, all of them ignored the impact of gradient discontinuity on NeuS, therefore additional supervision or tricks are introduced to stabilize training. Neuralangelo~\cite{li2023neuralangelo} proposes numerical gradient as an alternative to analytical gradient on iNGP hash grid, which alleviates the gradient discontinuity to a certain extent, but also brings the multiplied amount of the computational overhead.
As a comparison, our proposed gradient interpolation is able to eliminate gradient discontinuity without additional computational overhead. Cooperated with a general color network, e.g. hash grid, our method achieves efficient surface reconstruction.

\section{Preliminary}

NeuS~\cite{wang2021neus} uses the SDF for surface representation and formulates equivalent opacity in volume rendering with SDF value and its gradient for unbiased surface reconstruction. For a ray $\myp(t)= \myo+t\myv $ with $n$ segments, where the segment lengths are $\left\{\delta_i|i=1,...,n\right\}$ and the mid-point distances are $\left\{t_i|i=1,...,n\right\}$, the NeuS volume rendering compute the approximate pixel color of the ray as:
\begin{equation}
\label{eq:blending}
    \hat{\myC} = \sum_{i=1}^nT_i\alpha_i\myc_i,
\end{equation}
where $T_i=\prod_{j=1}^{i-1}(1-\alpha_j)$ is the accumulated transmittance, $\myc_i$ is the estimated color of segments, and $\alpha_i$ is discrete opacity values defined as:
\begin{equation}
    \label{eq:neus_vr}
    \alpha_i=\max\left(\frac{
    \Phi_s(f(\myp(t_i))-\frac{\delta_i}{2}\cos\theta_i) - \Phi_s(f(\myp(t_i))+\frac{\delta_i}{2}\cos\theta_i)
    }{\Phi_s(f(\myp(t_i))-\frac{\delta_i}{2}\cos\theta_i)},0\right).
\end{equation}
$\Phi_s(z)=(1+e^{-sz})^{-1}$ is the Sigmoid function, $f(\cdot)$ is the SDF network to be optimized, and $\cos\theta_i=\langle \myn_i, \myv \rangle$ is the cosine of the angle between the SDF gradient $\myn_i=\nabla_\myx f(\myx)|_{\myx=\myp(t_i)}$ and the ray direction $\myv$.

As shown in Eq.~\ref{eq:neus_vr}, both SDF values and their gradients participate in the forward pass of volume rendering. Therefore, when designing a volumetric representation for NeuS, properties of both the SDF value and its gradient (such as differentiability and continuity) need to be taken into consideration.

\begin{figure}[t]
  \centering
  \includegraphics[width=1\linewidth]{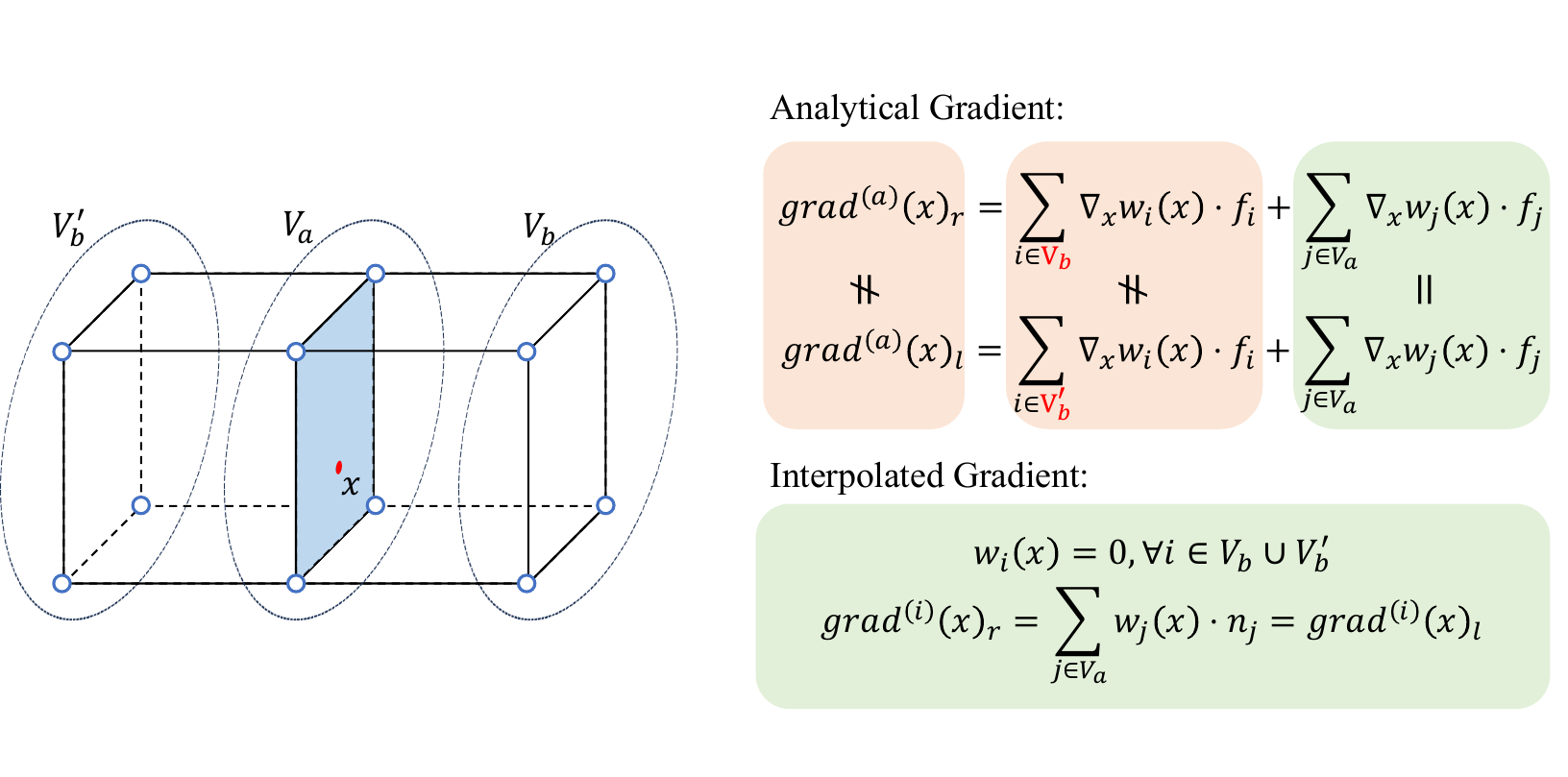}
  \caption{Gradient continuity on junction surface. A point $\myx$ lies on the junction surface of two adjacent cubes. $V_a$ is the set of points on the junction surface while $V_b$ and $V_b'$ are the sets of points outside the junction surface. As for \textit{analytical gradient} at $\myx$, the $grad^{(a)}_r$ from cube $V_a\bigcup V_b$ is not agree with the $grad^{(a)}_l$ from cube $V_a\bigcup V_b'$, showing discontinuity. As for \textit{interpolated gradient} at $\myx$, as the weight of $V_b$ and $V_b'$ is zero, the $grad^{(i)}_r$ is agree with $grad^{(i)}_l$, showing continuity.}
  \label{fig:continuity}
\end{figure}

\section{Approach}

We analyze the reasons for gradient discontinuity and solve it with interpolated gradients in Sec.~\ref{sec:ig}, describe efficient regularization methods on the SDF grid in Sec.~\ref{sec:regularization}, geometry-radiance disentangled architecture in Sec.~\ref{sec:architecture}, and optimization details in Sec.~\ref{sec:details}.

\subsection{SDF Gradient Interpolation}
\label{sec:ig}

Without loss of generality, We formulate the $d$-linear interpolation on the SDF grid and derive its derivative with respect to coordinate $\myx$. $\myc_i(\myx)$ is the $i$-th corner of the cube where $\myx$ is located. $f(\cdot)$ stands for the interpolated value and $f[\cdot]$ stands for directly queried vertex value. The $d$-linear interpolation is given by Eq.~\ref{eq:interpolation},
\begin{equation}
\label{eq:interpolation}
\begin{aligned}
    f(\myx) &= \sum_{i=1}^{2^d}w_i(\myx)f[\myc_i(\myx)] \\
    w_i(\myx)&=\prod_{j=1}^d (1 - |\myx - \myc_i(\myx)|_j),
\end{aligned}
\end{equation}
where $f(\myx)$ is the sum of $2^d$ vertex values, weighted by $w_i$. We define the analytical gradient of Eq.~\ref{eq:interpolation} with respect to $\myx$ as $grad^{(a)}(\myx)=\nabla_\myx f(\myx)$, which can be formulated as Eq.~\ref{eq:grada},
\begin{equation}
\label{eq:grada}
\begin{aligned}
    grad^{(a)}&(\myx) = \sum_{i=1}^{2^d}\nabla_\myx w_i(\myx)\cdot f[\myc_i(\myx)] \\
    \frac{\partial w_i(\myx)}{\partial \myx_k} = sign&(\myc_i(\myx)-\myx)_k \cdot \prod_{j\neq k} (1 - |\myx - \myc_i(\myx)|_j),
\end{aligned}
\end{equation}
where $grad^{(a)}(\myx)$ is the sum of $2^d$ vertex values, weighted by $\nabla_\myx w_i(\myx)\in \mathbb{R}^d$.

\begin{figure*}[t]
    \centering
    \begin{subfigure}[c]{0.48\linewidth}
        \centering
        \includegraphics[width=\linewidth]{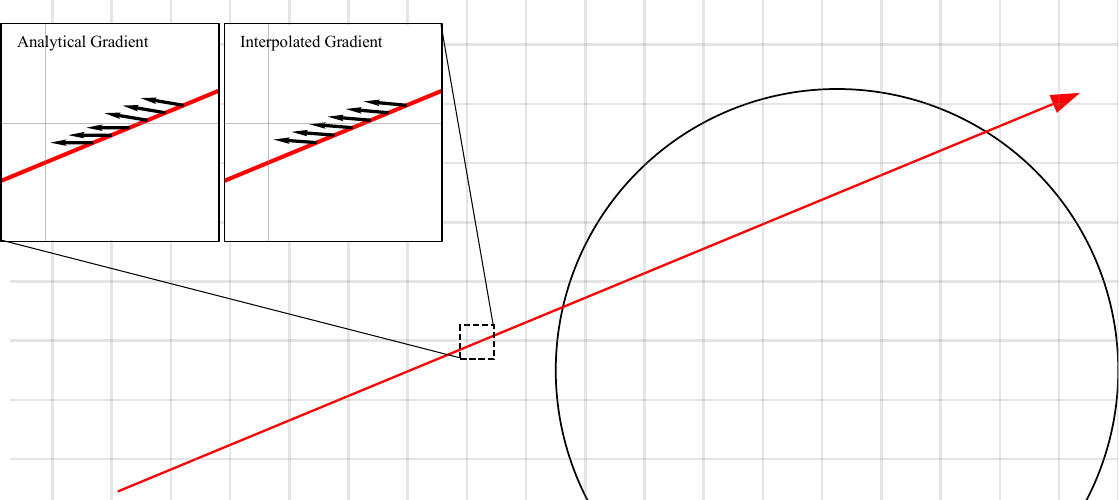}
     \end{subfigure}
     \hfill
     \begin{subfigure}[c]{0.48\linewidth}
        \centering
        \includegraphics[width=\linewidth]{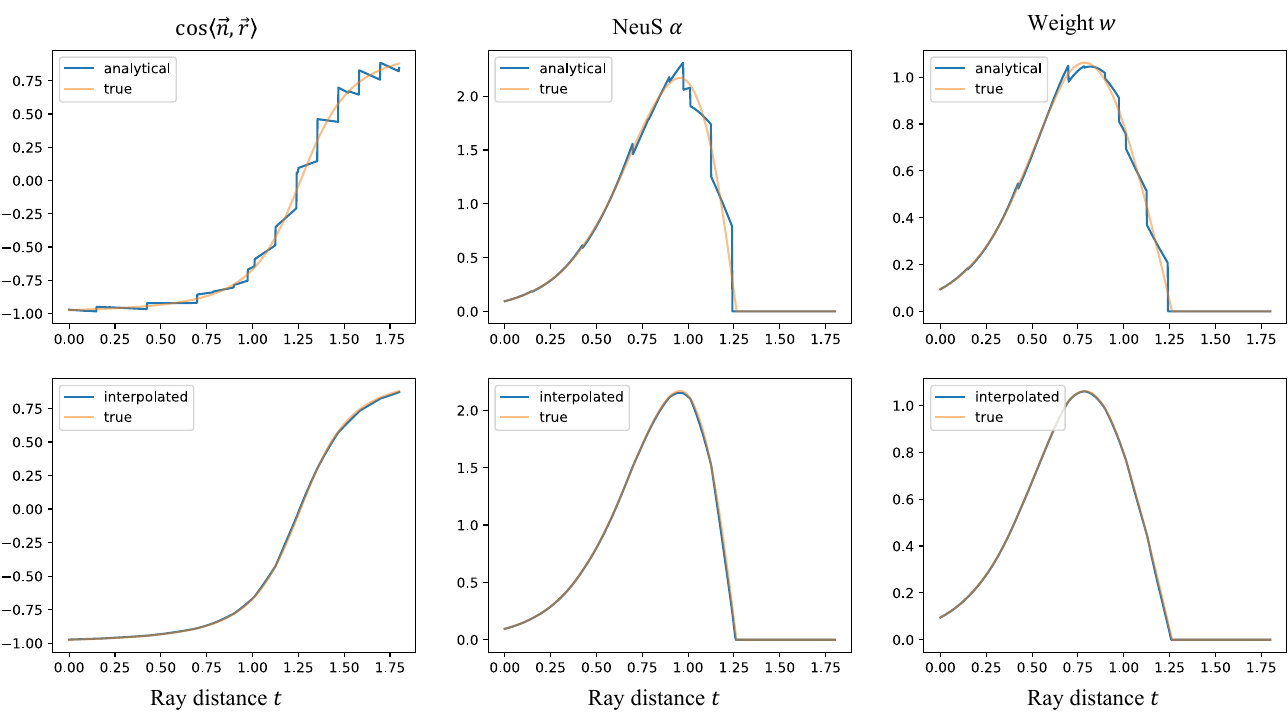}
     \end{subfigure}
    \caption{An 2D example to illustrate the continuity of the analytical and interpolated gradient. In this example, a ray is marching to the curved surface. The ground truth SDF values are stored on the vertices of the dense grid. The analytical gradient is computed with Eq.~\ref{eq:grada} and the interpolated gradient is computed with Eq.~\ref{eq:gradi1}. The discontinuity of the analytical gradient can be observed at the junction surface (left). Such discontinuity causes glitches on $cos\langle \myn, \myv\rangle$, opacity $\alpha$, and volume rendering weights $w$ (right). After applying an interpolated gradient, the estimated gradient gets close to ground truth and the glitches are eliminated.}
    \label{fig:example}
\end{figure*}

Then, we analyse the continuity of $grad^{(a)}(\myx)$ on the junction surface of adjacent cubes. Consider a point $\myx$ lies on the junction surface. Here we assume the junction surface is the little end of the first dimension, i.e. $\myx_1 = \lfloor \myx \rfloor_1$. Next, we split the corners into 2 groups: $V_a = \left\{\myc_i(\myx)|\myc_i(\myx)_1 = \myx_1\right\}$ and $V_b=\left\{\myc_i(\myx)|\myc_i(\myx)_1 = \myx_1+1\right\}$. $V_a$ are on the junction surface while $V_b$ are not. $grad^{(a)}(\myx)$ is the weighted sum of the values of vertices in $V_a$ and $V_b$. The weights of vertices in $V_b$, i.e. $\left\{\nabla_\myx w_i(\myx)|\myc_i(\myx) \in V_b\right\}$, given by Eq.~\ref{eq:grada_w},
\begin{equation}
\label{eq:grada_w}
\begin{aligned}
    \nabla_\myx w_i(\myx) = \left[\prod_{j=2}^d(1 - |\myx - \myc_i(\myx)|_j),\textbf{0}^{(d-1)}\right]^T,\\
    where\quad \myc_i(\myx) \in V_b
\end{aligned}
\end{equation}
is non-zero, thus leading to disagreement between two adjacent cubes. Therefore, $grad^{(a)}(\myx)$ is not continuous on the junction surface. A more intuitive illustration can be found in Fig.~\ref{fig:continuity}.

According to the NeuS volume rendering scheme in Eq.~\ref{eq:neus_vr}, the discontinuity of gradient results in the fluctuation of opacity $\alpha_i$ and further affects the volume rendering. An example in 2D is shown in Fig.~\ref{fig:example}, where a ray hits a curved surface. Discontinuous gradients lead to fluctuations in the curves of cosine, opacity and weight.

Based on the observations above, gradient continuity is critical to NeuS optimization. We propose to replace the analytical gradient with an interpolated gradient to ensure gradient continuity. Specifically, analogous to the interpolation of SDF values, we perform the interpolation of the gradient, which is estimated through central finite differences, so that the gradient becomes continuous. In 3D space, the estimated vertex gradient $\myn[x,y,z]$ is given by Eq.~\ref{eq:gradi}:
\begin{equation}
\label{eq:gradi}
\begin{aligned}
    \frac{\partial f[x,y,z]}{\partial x} &= (f[x+1,y,z] - f[x-1,y,z]) / 2\epsilon, \\
    \frac{\partial f[x,y,z]}{\partial y} &= (f[x,y+1,z] - f[x,y-1,z]) / 2\epsilon, \\
    \frac{\partial f[x,y,z]}{\partial z} &= (f[x,y,z+1] - f[x,y,z-1]) / 2\epsilon, \\
    \myn[x,y,z] &= \left(\frac{\partial f[x,y,z]}{\partial x},\frac{\partial f[x,y,z]}{\partial y},\frac{\partial f[x,y,z]}{\partial z}\right).
\end{aligned}
\end{equation}
Then, the interpolated gradient $grad^{(i)}(\myx)$ is defined by Eq.~\ref{eq:gradi1},
\begin{equation}
\label{eq:gradi1}
    grad^{(i)}(\myx) = \sum_{i=1}^{2^d}w_i(\myx)\myn[\myc_i(\myx)].
\end{equation}

Similarly, we analyse the continuity of $grad^{(i)}(\myx)$ in Eq.~\ref{eq:gradi1}. The weights of vertices in $V_b$, i.e. $\left\{w_i(\myx)|\myc_i(\myx) \in V_b\right\}$ is given by:
\begin{equation}
\begin{aligned}
    w_i(\myx)=\prod_{j=1}^d (1 - |\myx - \myc_i(\myx)|_j)=\textbf{0},\\
    where\quad \myc_i(\myx)\in V_b
\end{aligned}
\end{equation}
which means that the interpolated gradient $grad^{(i)}(\myx)$ is continuous at the junction surface. As shown in Fig.~\ref{fig:example}, the interpolated gradient eliminates the fluctuations of volume rendering weights.

\subsection{Efficient SDF Regularization}
\label{sec:regularization}

In VoxNeuS, we apply Eikonal loss~\cite{gropp2020implicit} and curvature loss~\cite{li2023neuralangelo} to SDF dense grids. The regularization is not applied to the interpolated points but the vertices of the grid. This design achieves the same regularization effect while avoiding interpolation, thereby improving efficiency.




Specifically, the vertices set for regularization is given by Eq.\ref{rq:reg_set}:
\begin{equation}
\label{rq:reg_set}
    V_R=\left\{\myc_i(\myx)|i=1,...,2^d;\myx\in \mathcal{B}\right\}
\end{equation}
where $\mathcal{B}$ is the sampled points on the batch of rays for volume rendering. The vertices in $V_R$ are uniformly regularized.

Current learning framework, like PyTorch, supports automatic differentiation. However, when handling the regularization term on vertices set $V_R$, frequent unstructured memory access makes it a bottleneck. To accelerate this computation, we take advantage of the explicit representation of SDF values to directly solve for the gradient of the regularization term. These manually solved gradients will be added to the gradient obtained by automatic differentiation, i.e. $\boldsymbol{\mathcal{G}}\leftarrow \boldsymbol{\mathcal{G}} + w_{eik}\boldsymbol{\mathcal{G}}_{eik} + w_{curv}\boldsymbol{\mathcal{G}}_{curv}$.

The Eikonal loss function makes the learned field satisfy the SDF property~\cite{barrett2020implicit}. Its formulation and derivative with respect to finite difference $n[x]$ are given by Eq.~\ref{eq:eikonal}:
\begin{equation}
\label{eq:eikonal}
\begin{aligned}
    \mathcal{L}_{eik} = \frac{1}{|V_R|}\sum_{\myx\in V_R}(||\myn[\myx]||_2 - 1)^2 \\
    \frac{\partial \mathcal{L}_{eik}}{\partial \myn[\myx]} = \frac{1}{|V_R|}(1-||\myn[\myx]||_2^{-1}).
\end{aligned}
\end{equation}
The derivative of $\myn[\myx]$ with respect to $\myx$'s neighbourhood SDF values $f[U(\myx)]$ can be derived from Eq.~\ref{eq:gradi}. Combining it with Eq.~\ref{eq:eikonal}, the $\mathcal{G}_{eik}$ is formulated as:
\begin{equation}
    \boldsymbol{\mathcal{G}}_{eik} = \sum_{\myx\in V_R} \frac{\partial \mathcal{L}_{eik}}{\partial \myn[\myx]} \cdot \frac{\partial \myn[\myx]}{\partial f[U(\myx)]}.
\end{equation}

We penalize the mean squared curvature of SDF to encourage smoothness. The curvature is computed from discrete Laplacian, as given by Eq.~\ref{eq:laplacian} ($d=3$).
\begin{equation}
\label{eq:laplacian}
    \nabla^2f[x,y,z] =
    \frac{1}{\epsilon^2}
    \left[\begin{aligned}
        f[x+1,y,z] + f[x-1,y,z] - 2f[x,y,z]\\
        f[x,y+1,z] + f[x,y-1,z] - 2f[x,y,z]\\
        f[x,y,z+1] + f[x,y,z-1] - 2f[x,y,z]\\
    \end{aligned}\right]
\end{equation}
The curvature loss and its derivative are defined as Eq.~\ref{eq:curvature},
\begin{equation}
\label{eq:curvature}
\begin{aligned}
    \mathcal{L}_{curv} =& \frac{1}{|V_R|}\sum_{\myx\in V_R}||\nabla^2f[\myx]||^2_2\\
    \boldsymbol{\mathcal{G}}_{curv} = \frac{2}{|V_R|}&\sum_{\myx\in V_R}\nabla^2f[\myx]\cdot \frac{\partial \nabla^2f[\myx]}{\partial f[U(\myx)]}.
\end{aligned}
\end{equation}

While larger curvature weight $w_{curv}$ encourages smoothness, it hinders the initial convergence and later detail optimization~\cite{li2023neuralangelo}. Therefore dedicated scheduling is applied: $w_{curv}$ is a small constant at first, linearly increases at the following stage, and exponentially decreases at the final stage.


\begin{figure}[t]
    \centering
    \includegraphics[width=\linewidth]{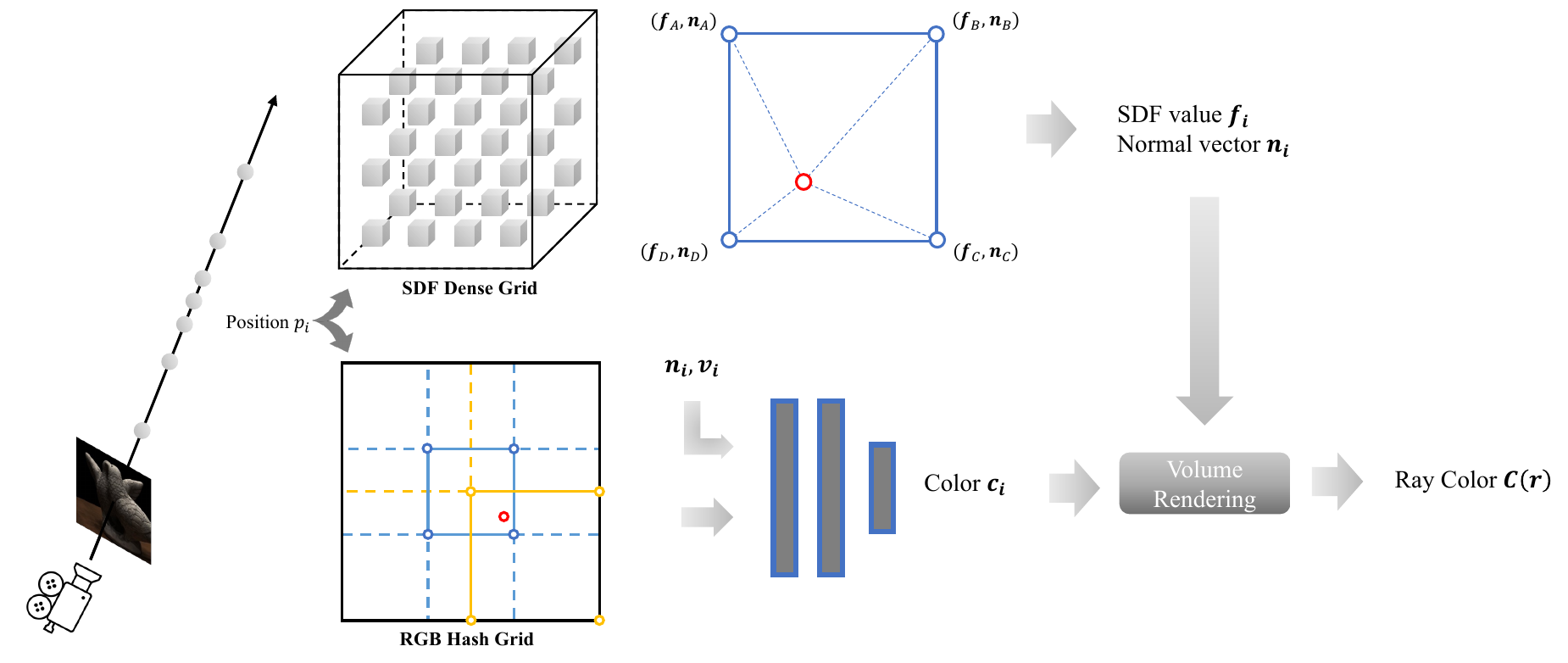}
    \caption{Network architecture overview.}
    \label{fig:main}
\end{figure}

\subsection{Network Architecture}
\label{sec:architecture}



\begin{figure*}[t]
    \includegraphics[width=\linewidth]{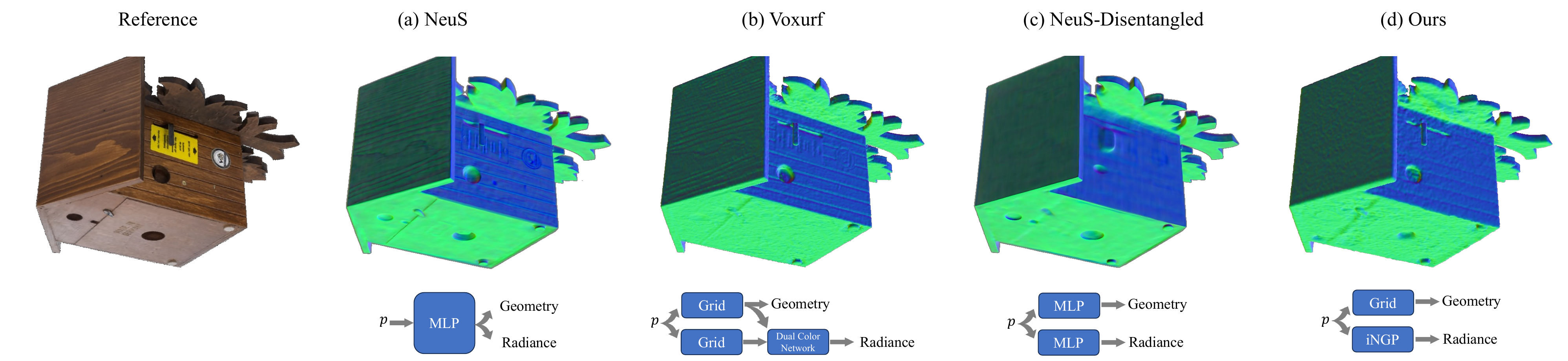}
    \caption{An illustration of geometry-radiance entanglement in volume rendering. In this example, the woodgrain roof and side stickers should be reconstructed as a plane. (a) NeuS uses one MLP to parameterize both geometry and radiance, causing geometry deformation when optimizing radiance. (b) Voxurf uses separate models to parameterize geometry and radiance. However, dual color networks use local geometry to predict radiance, resulting in two sets of parameters re-entangled. Using color cues to assist geometry optimization can improve reconstruction quality to some extent, but can also lead to artifacts in this example. (c) We modify NeuS to use independent models to parameterize geometry and color. The roof and stickers are reconstructed correctly, but the reconstruction quality is degraded. (d) Our VoxNeuS, where geometry is explicitly represented with a dense SDF grid and radiance are parameterized with an iNGP hash grid, correctly reconstructed the textured plane and improved the quality and efficiency.}
    \label{fig:RG-disentanglement}
\end{figure*}

In volume rendering, geometry and radiance are handled in two independent passes, merged when blending the sampled points (Eq.~\ref{eq:blending}). However, general implementations adopt the same network for the prediction of both radiance and geometry. Such an entanglement lets the optimization of radiance pass affect the geometry. The problem appears when rendering a textured plane, as shown in Fig.~\ref{fig:RG-disentanglement}, where the plane is slightly deformed to fit the texture. Though Voxurf~\cite{wu2022voxurf} uses two separate DVGO~\cite{sun2022direct} grids as its architecture, the feature branch of dual color network predicts the point color based on local SDF values, causing the geometry to be optimized through the color prediction pass.

Based on the observations above, we adopt a geometry-radiance disentangled architecture, where the geometry is explicitly represented with an SDF dense grid and the radiance is parameterized with an iNGP hash grid, as shown in Fig.~\ref{fig:main}. Following IDR~\cite{yariv2020multiview}, the normal is passed to the iNGP's shallow MLP, which leads to slight optimization of geometry through radiance pass. However, no artifacts in the textured plane are observed.

\subsection{Optimization Details}
\label{sec:details}



We describe optimization details, and implementations to improve efficiency in this section.

In VoxNeuS, we leverage a progressive super-resolution of the SDF grid to balance the coarse shape and fine detail. A small SDF dense grid is used at the beginning to initialize a coarse geometry and scaled up during training to capture details. The final resolution is $320^3$. Following~\cite{yariv2023bakedsdf}, we manually schedule the $s$ value in $\Phi_s$ (Eq.~\ref{eq:neus_vr}) for more accurate surface location. Specifically, the gradient $\nabla_s\mathcal{L}$ is amplified by $k$ times when negative (towards convergence), as formulated in Eq.~\ref{eq:anneals},
\begin{equation}
\label{eq:anneals}
    \nabla_s\mathcal{L} \leftarrow \left\{
    \begin{aligned}
        k \nabla_s\mathcal{L} \quad &\nabla_s\mathcal{L} < 0\\
        \nabla_s\mathcal{L} \quad &\nabla_s\mathcal{L} >= 0,
    \end{aligned}
    \right.
\end{equation}
where we set $k=5$ for all our experiments.

As the regularization term is handled by explicit gradient accumulation (Sec.~\ref{sec:regularization}), we only perform the RGB and optional mask supervision, as shown in Eq.~\ref{eq:loss},
\begin{equation}
\label{eq:loss}
    \mathcal{L} = \mathcal{L}_{RGB} + w_{mask}\mathcal{L}_{mask}.
\end{equation}
Between the back-propagation of $\mathcal{L}$ and gradient descent, the regularization gradients~(Eq.\ref{eq:eikonal} \& \ref{eq:curvature}) need to be added. After training, an optional Gaussian filtering can be applied for smoothness~\cite{wu2022voxurf}.

To further improve the computational and memory efficiency, critical operators are implemented with CUDA. For native PyTorch-based gradient interpolation, on the one hand, additional memory is required to save intermediate results for back-propagation, and on the other hand, non-local memory access results in a longer computation time. Therefore, we use CUDA to achieve local memory access, and use re-computation in backward pass to avoid the memory overhead of intermediate results. The explicit gradient computation of Eq.~\ref{eq:eikonal} \& Eq.~\ref{eq:curvature} is also implemented with CUDA for local memory access. Besides, the gradient descent introduces the multiply and add operation on the entire dense grid, which is unignorable when the resolution becomes higher. Following iNGP~\cite{muller2022instant}, we utilize the sparsity of gradient and only update the SDF values whose gradient is non-zero. Although this strategy violates Adam's assumption, it does not cause a loss of precision.

\section{Experiments}

\subsection{Experiment Setup}

We evaluate our method on DTU~\cite{jensen2014large} dataset quantitatively and qualitatively, and further show qualitative results on several challenging scenes from the BlendedMVS~\cite{yao2020blendedmvs}.  We include several baselines for comparison: 1) Colmap~\cite{schoenberger2016sfm}; 2) IDR~\cite{yariv2020multiview}; 3) NeuS~\cite{wang2021neus}; 4) VolSDF~\cite{yariv2021volume}; 5) NeRF~\cite{mildenhall2021nerf}; 6) Voxurf~\cite{wu2022voxurf}; 7) NeuS2~\cite{wang2023neus2}; 8) 2DGS~\cite{huang20242d}. Among them, only 2) and 7) require foreground masks.

We train our models for 40k steps with ray batch size of 2048. The resolution of dense SDF grid is set to $96^3$ at beginning, scaling to $160^3$ after 10k steps, and scaling to $320^2$ after next 20k steps. We set the Eikonal weight $w_{eik}=10^{-2}$ for first 11k steps, linearly decrease to $10^{-3}$ for the next 10k steps. We set the curvature weight $w_{curv}=10^{-8}$ for the first 11k steps, linearly increase to $5\times 10^{-6}$ for the next 10k steps, and exponentially decrease to $5\times 10^{-7}$ during the remaining steps.

\subsection{Comparative Study}

\begin{figure*}[t]
    \centering

    \begin{subfigure}[b]{0.16\linewidth}\centering\includegraphics[width=\linewidth]{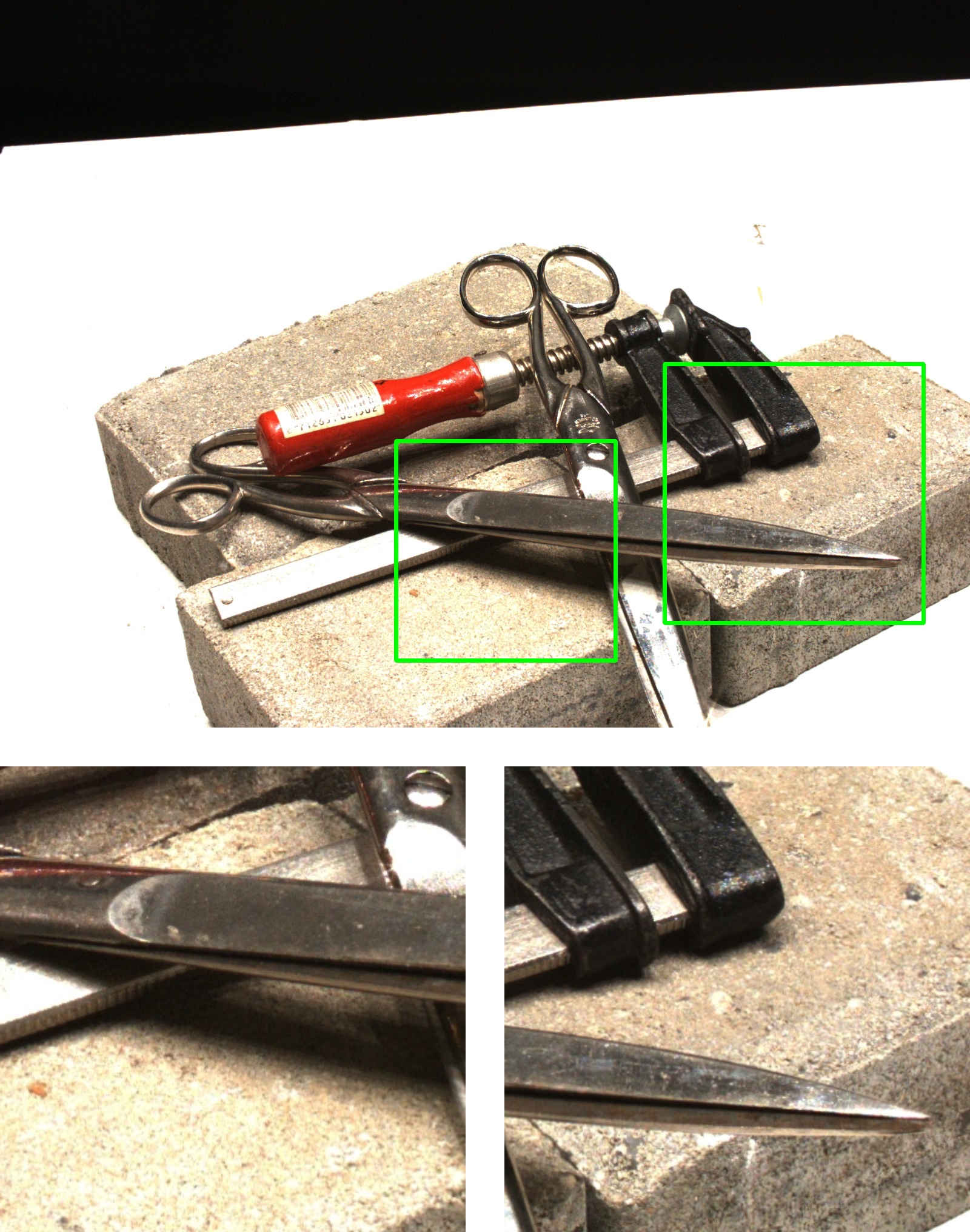}\vspace{2mm}\end{subfigure}
    \begin{subfigure}[b]{0.16\linewidth}\centering\includegraphics[width=\linewidth]{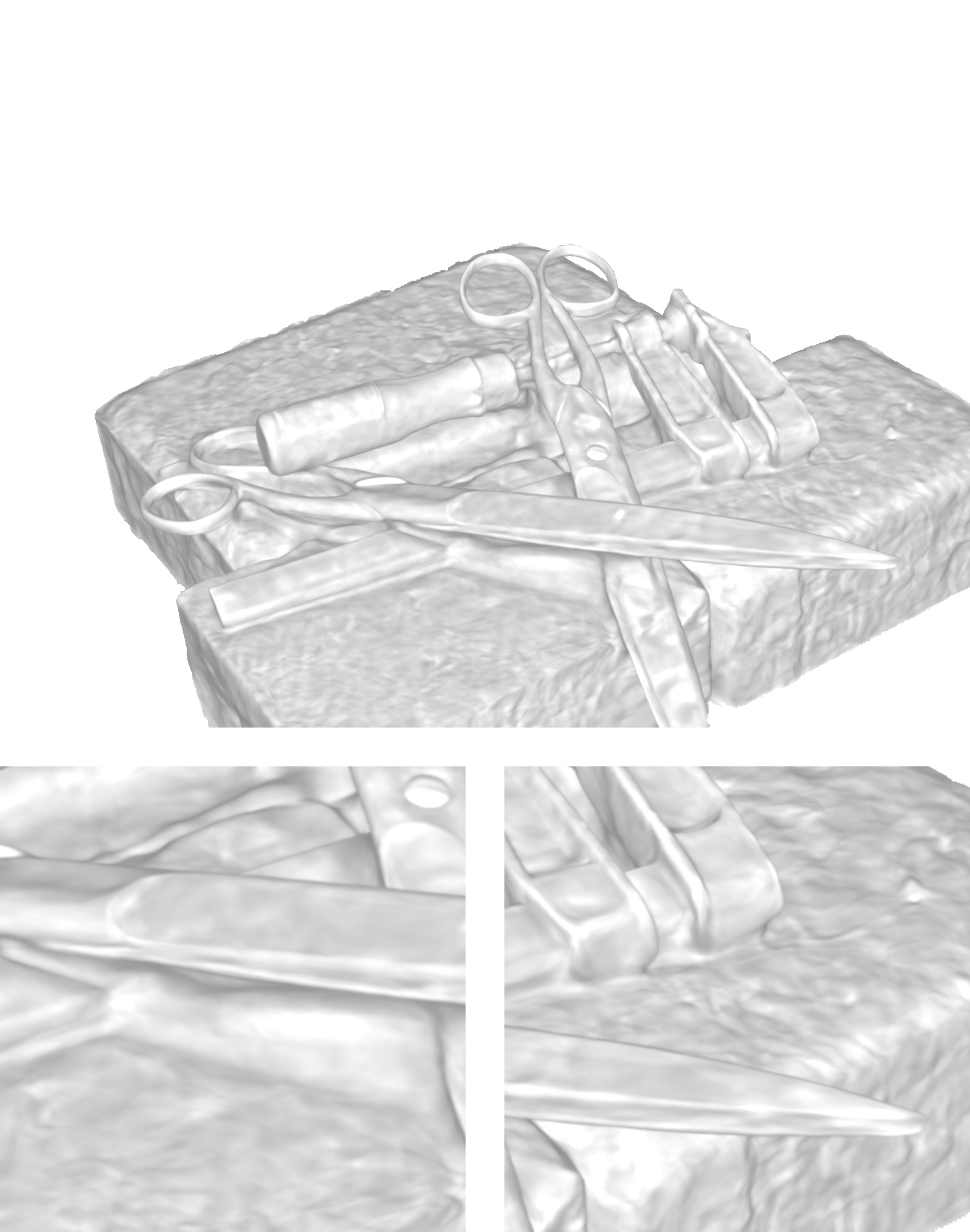}\vspace{2mm}\end{subfigure}
    \begin{subfigure}[b]{0.16\linewidth}\centering\includegraphics[width=\linewidth]{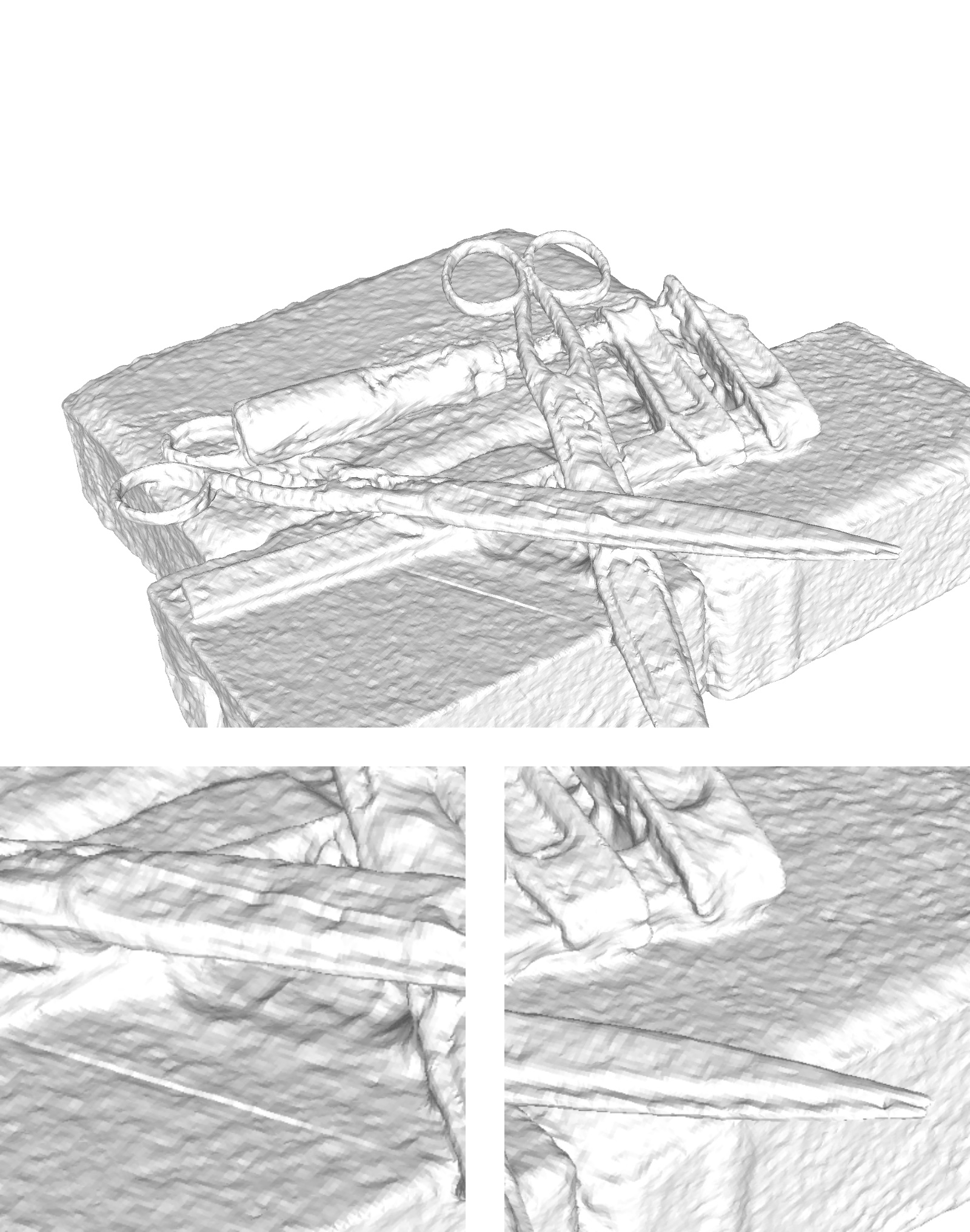}\vspace{2mm}\end{subfigure}
    \begin{subfigure}[b]{0.16\linewidth}\centering\includegraphics[width=\linewidth]{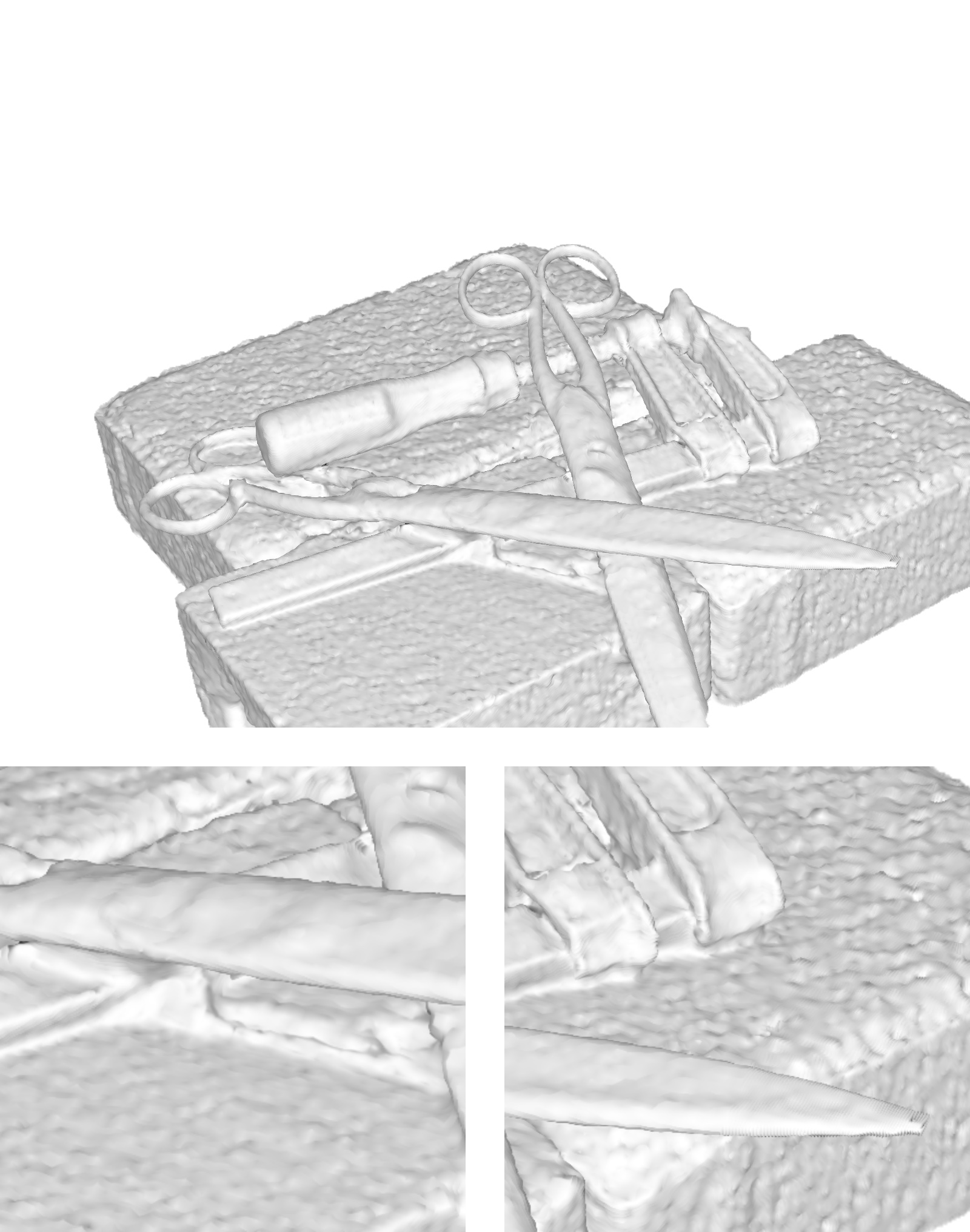}\vspace{2mm}\end{subfigure}
    \begin{subfigure}[b]{0.16\linewidth}\centering\includegraphics[width=\linewidth]{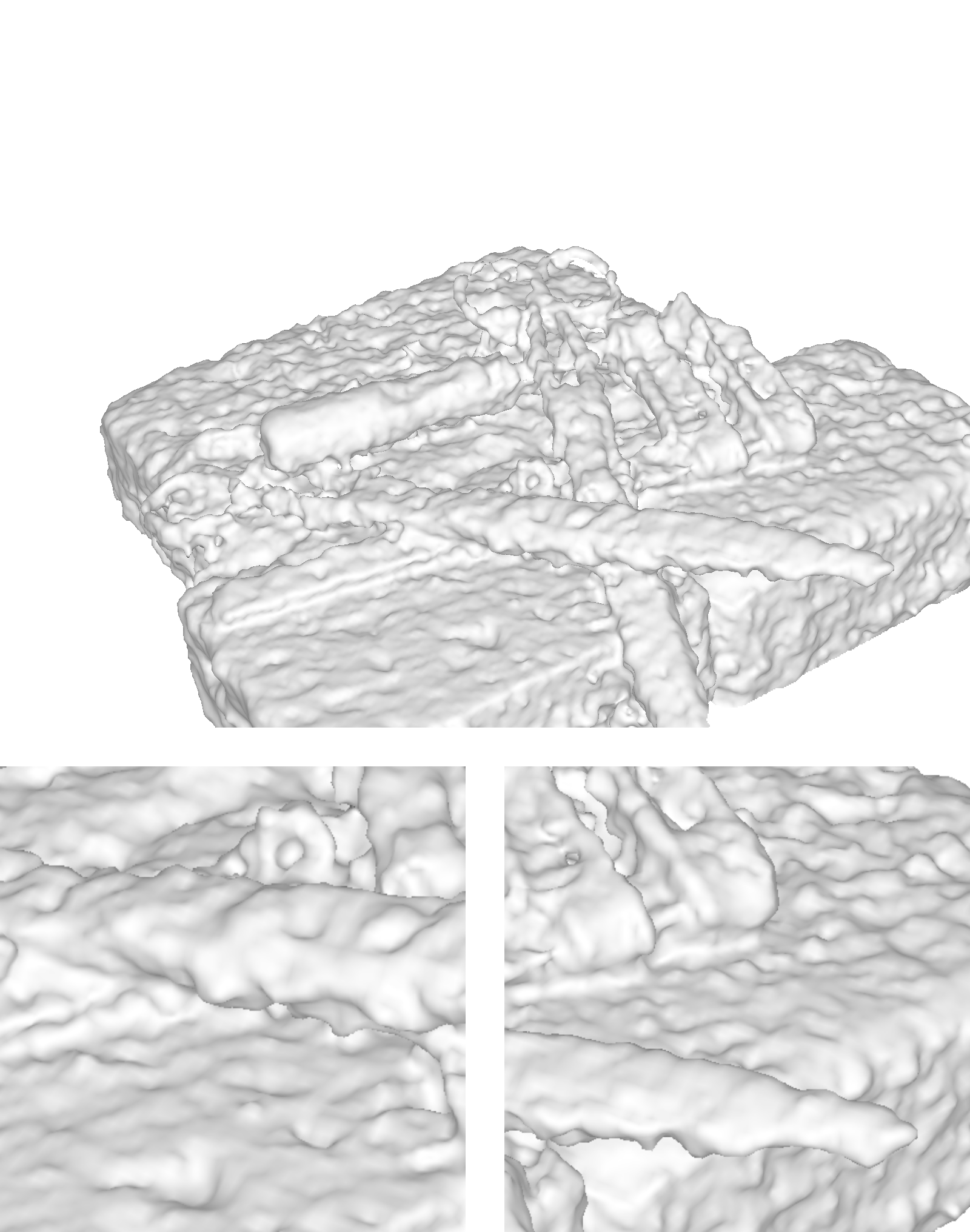}\vspace{2mm}\end{subfigure}
    \begin{subfigure}[b]{0.16\linewidth}\centering\includegraphics[width=\linewidth]{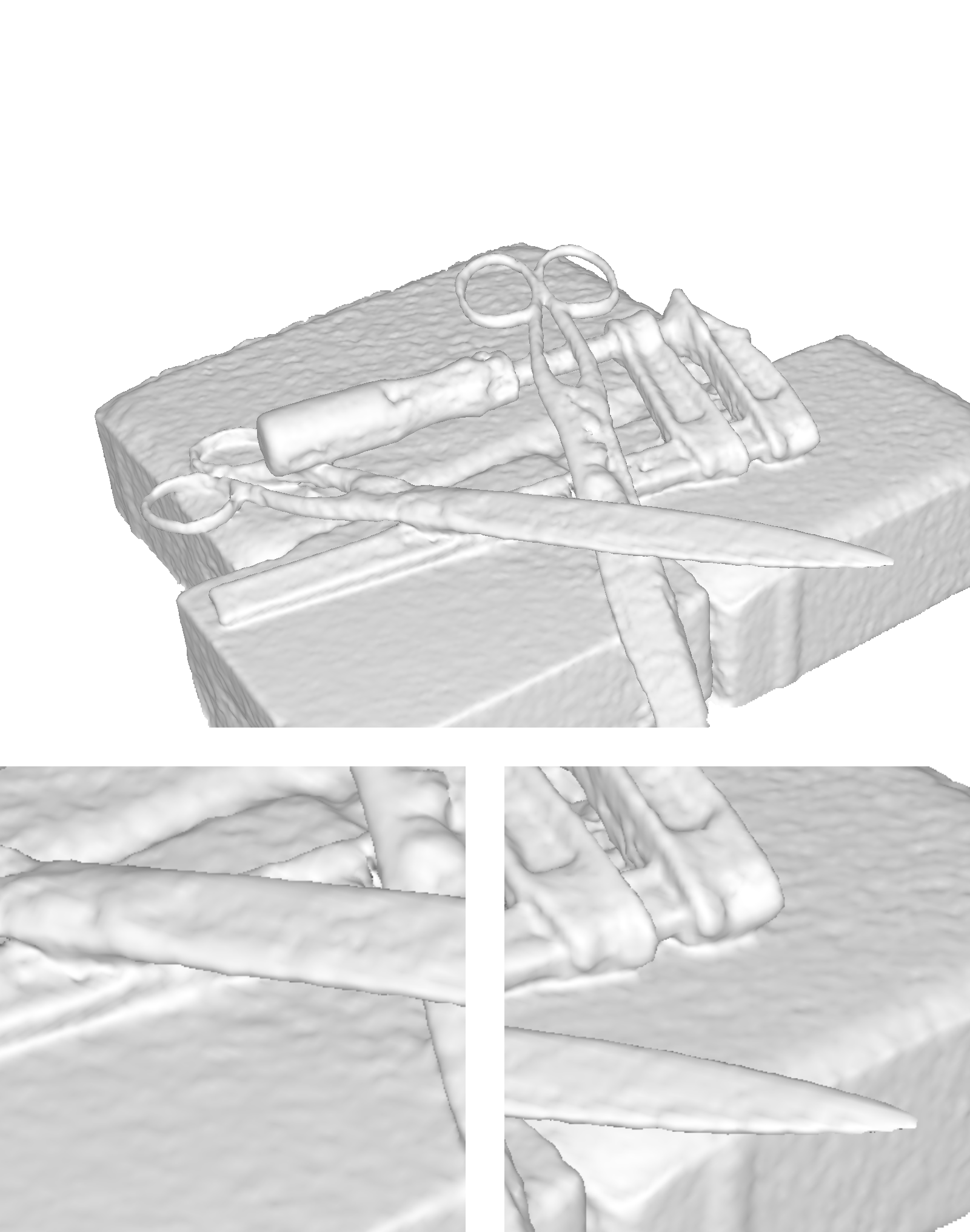}\vspace{2mm}\end{subfigure}

    \begin{subfigure}[b]{0.16\linewidth}\centering\includegraphics[width=\linewidth]{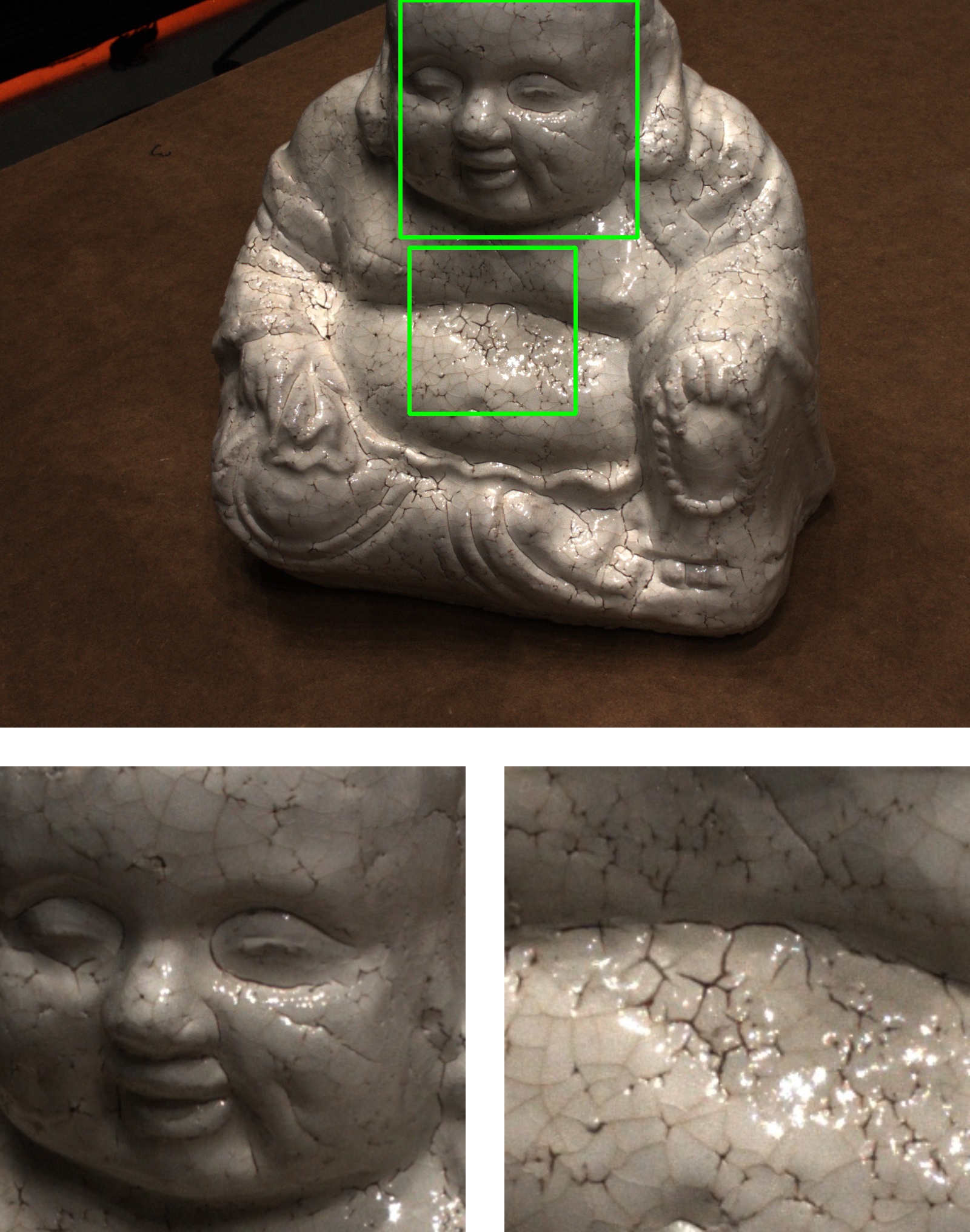}\vspace{2mm}\end{subfigure}
    \begin{subfigure}[b]{0.16\linewidth}\centering\includegraphics[width=\linewidth]{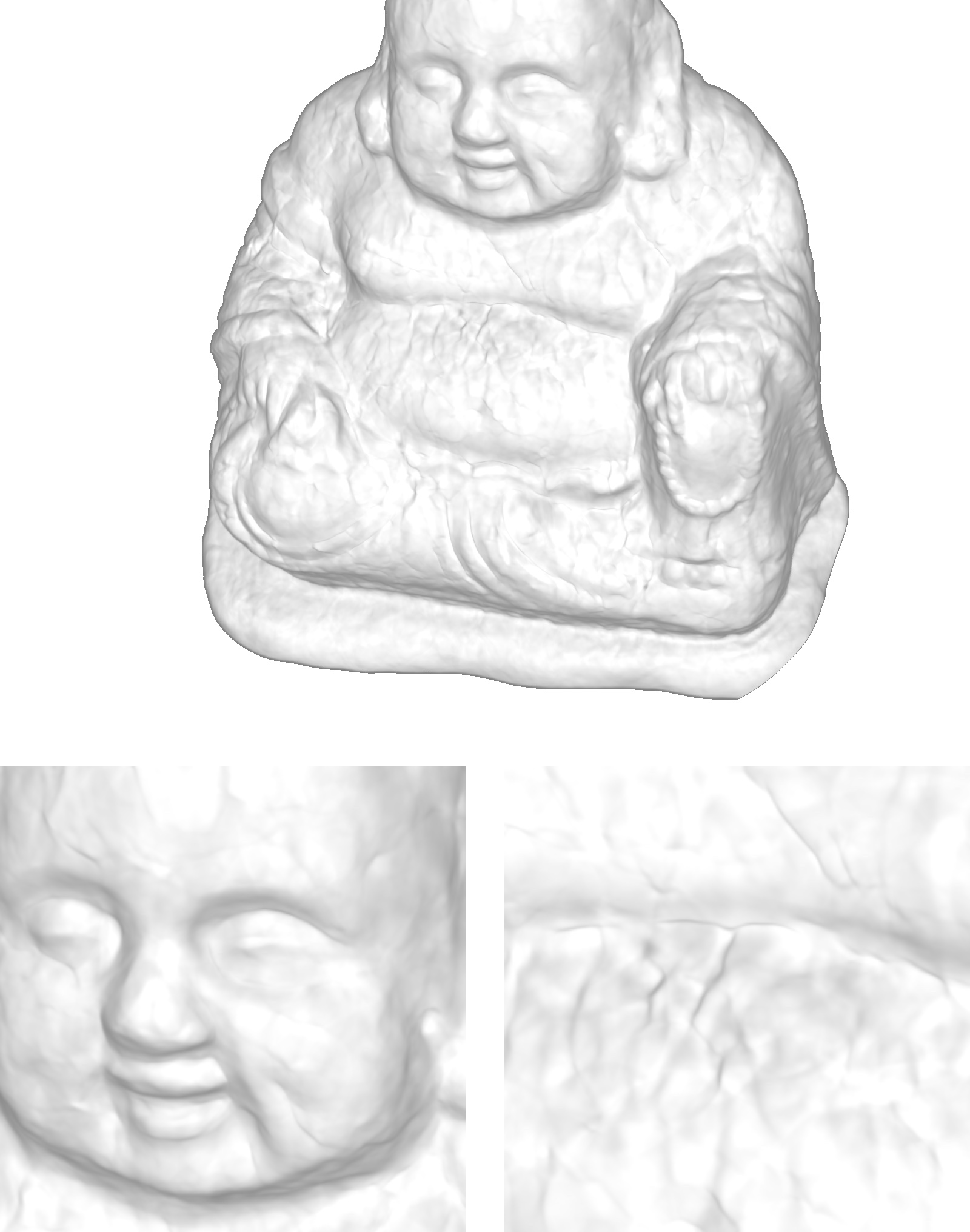}\vspace{2mm}\end{subfigure}
    \begin{subfigure}[b]{0.16\linewidth}\centering\includegraphics[width=\linewidth]{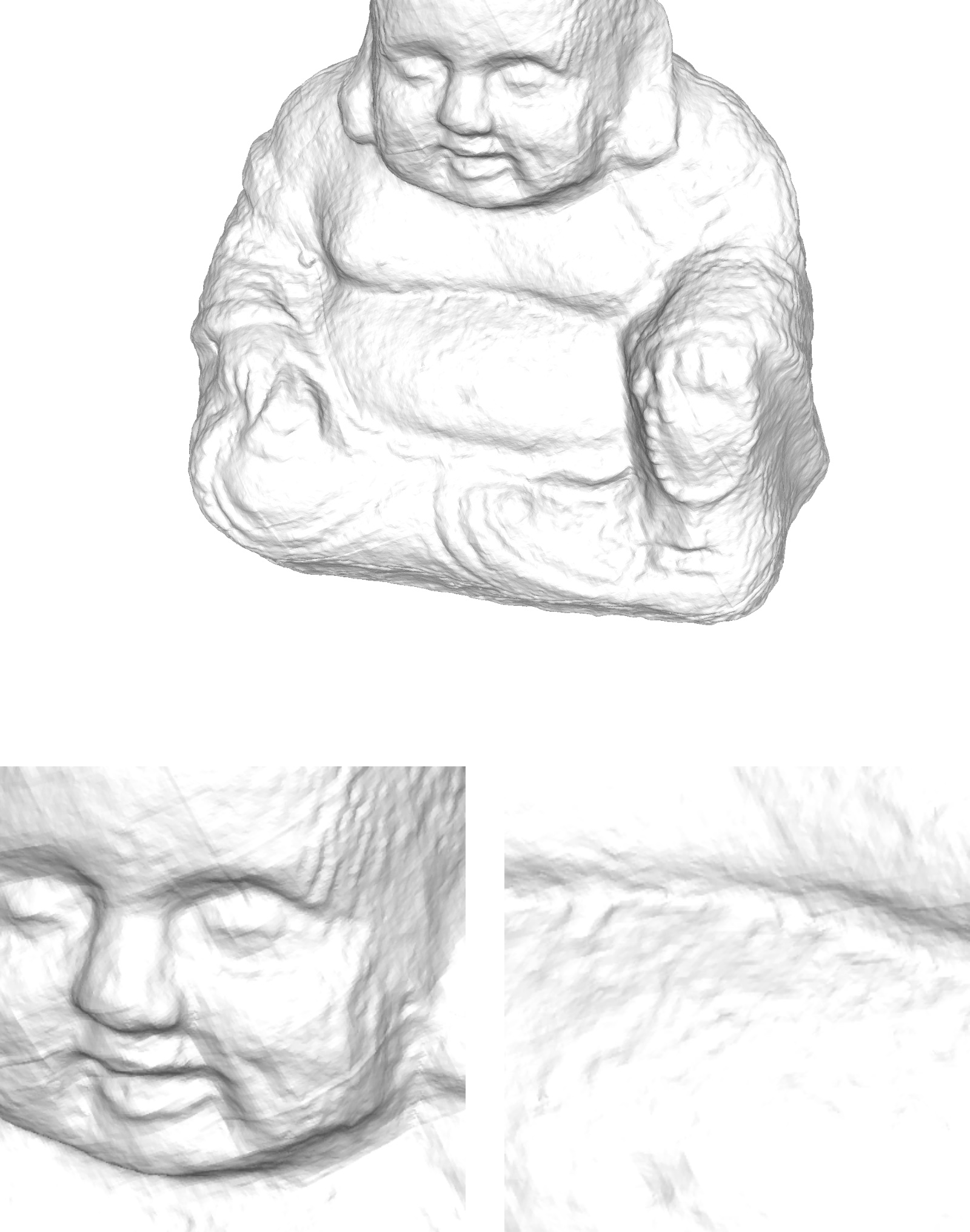}\vspace{2mm}\end{subfigure}
    \begin{subfigure}[b]{0.16\linewidth}\centering\includegraphics[width=\linewidth]{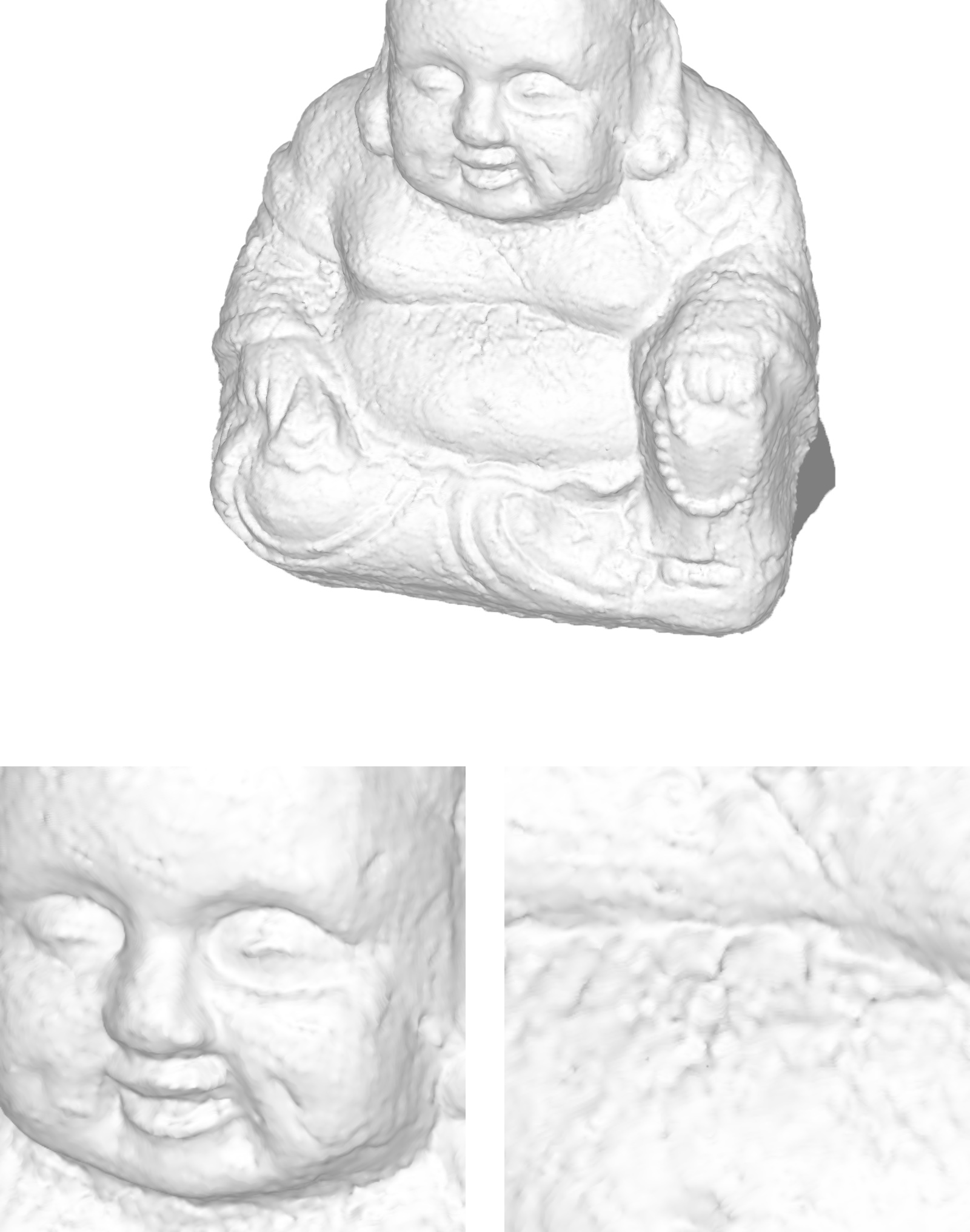}\vspace{2mm}\end{subfigure}
    \begin{subfigure}[b]{0.16\linewidth}\centering\includegraphics[width=\linewidth]{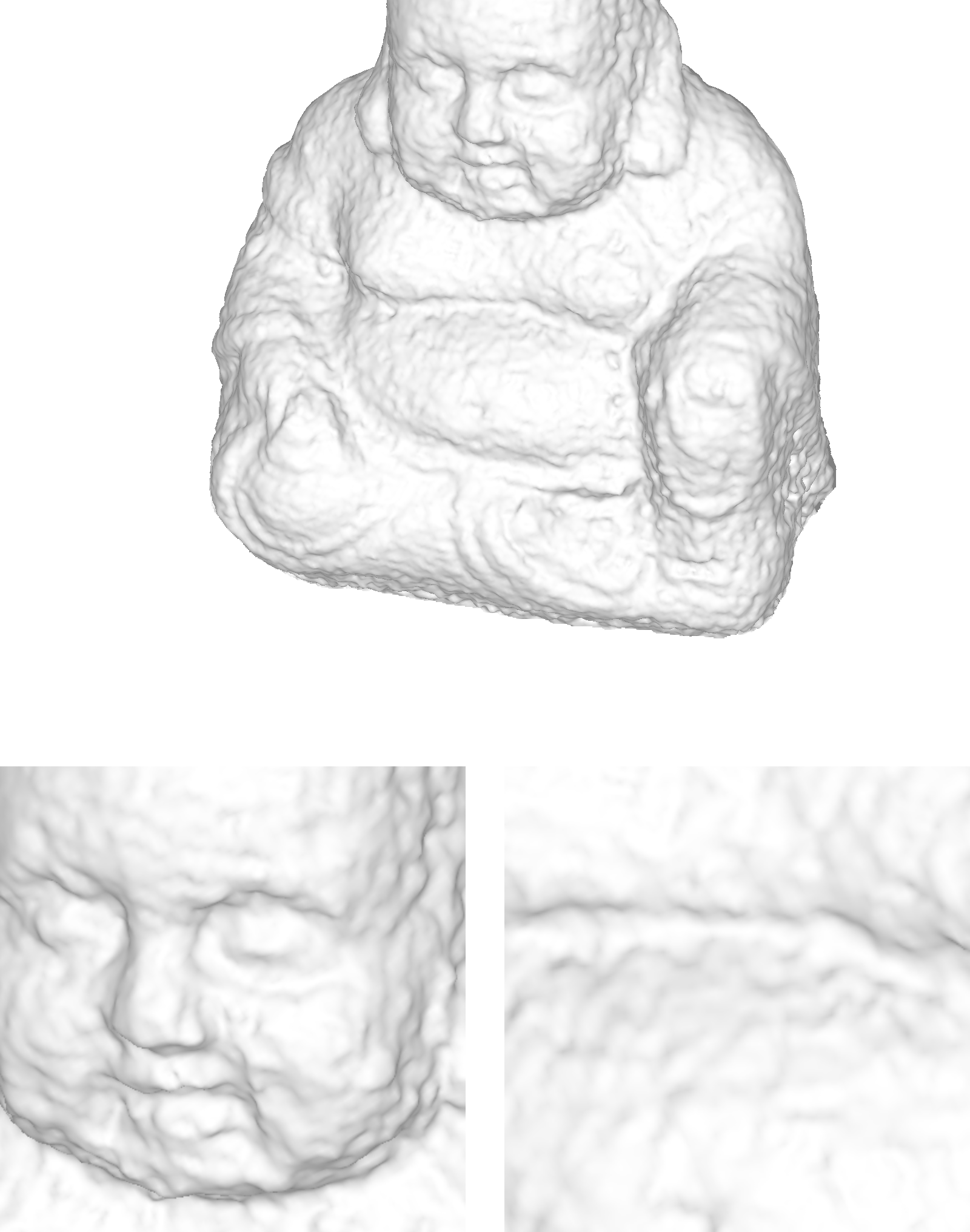}\vspace{2mm}\end{subfigure}
    \begin{subfigure}[b]{0.16\linewidth}\centering\includegraphics[width=\linewidth]{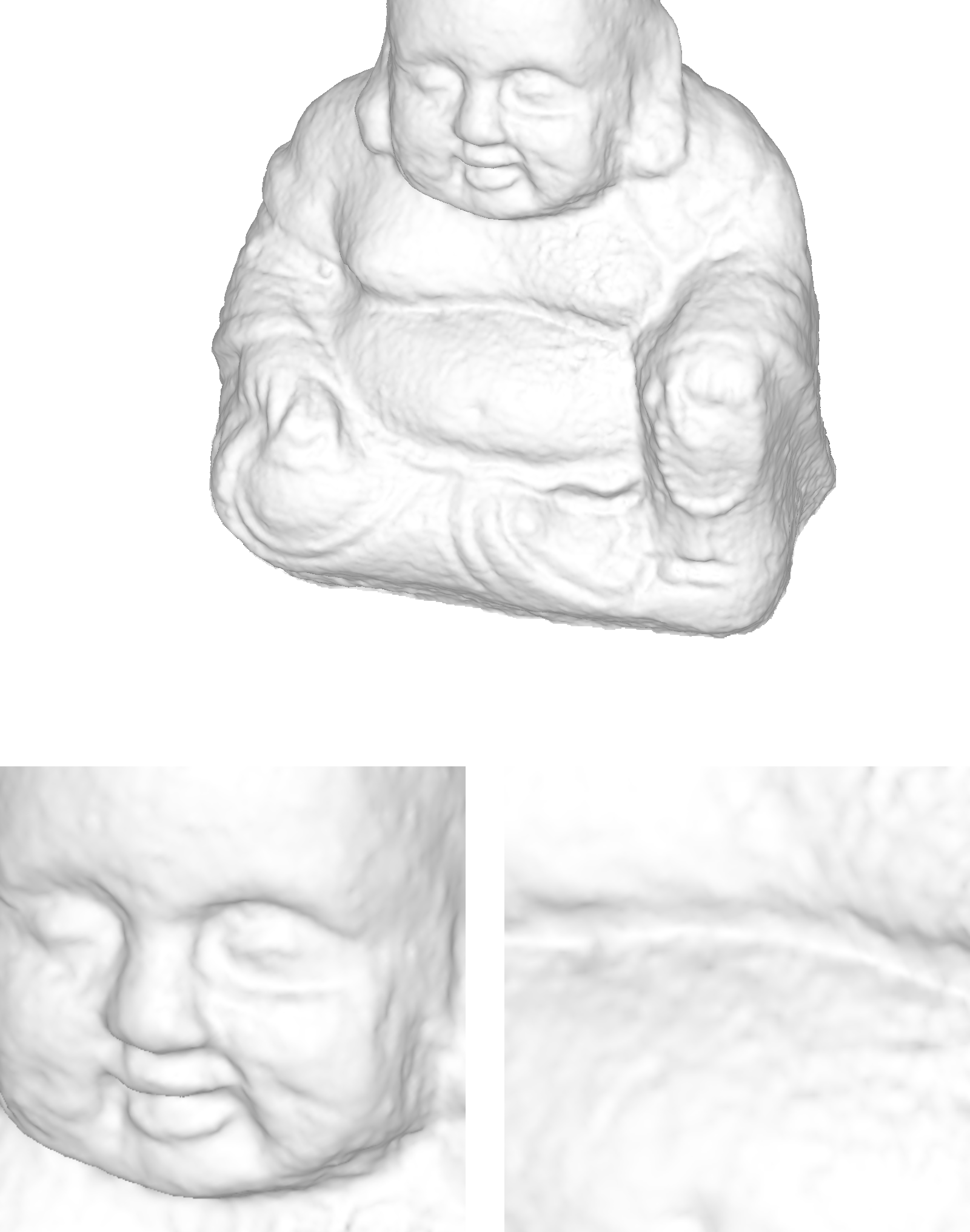}\vspace{2mm}\end{subfigure}

    \begin{subfigure}[b]{0.16\linewidth}\centering\includegraphics[width=\linewidth]{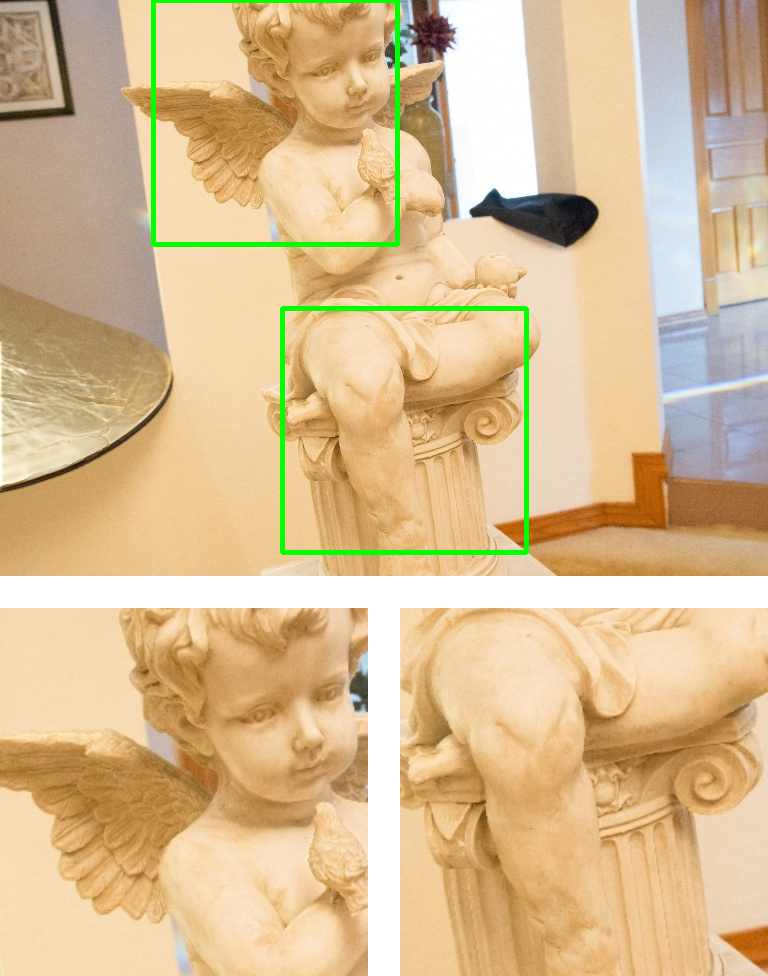}\vspace{2mm}\end{subfigure}
    \begin{subfigure}[b]{0.16\linewidth}\centering\includegraphics[width=\linewidth]{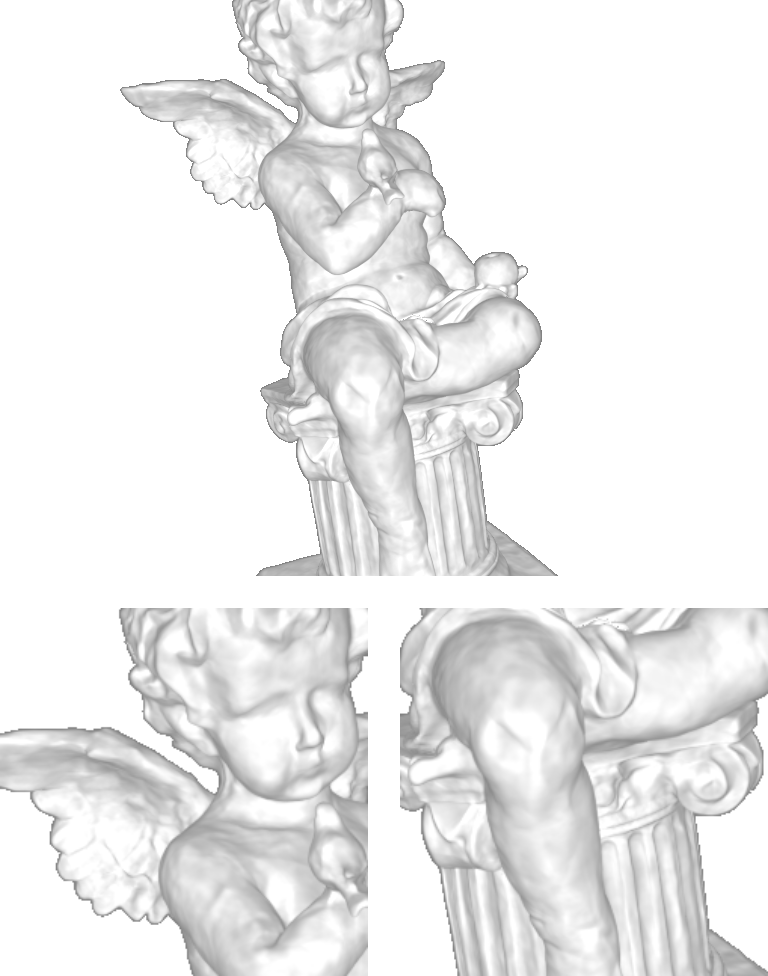}\vspace{2mm}\end{subfigure}
    \begin{subfigure}[b]{0.16\linewidth}\centering\includegraphics[width=\linewidth]{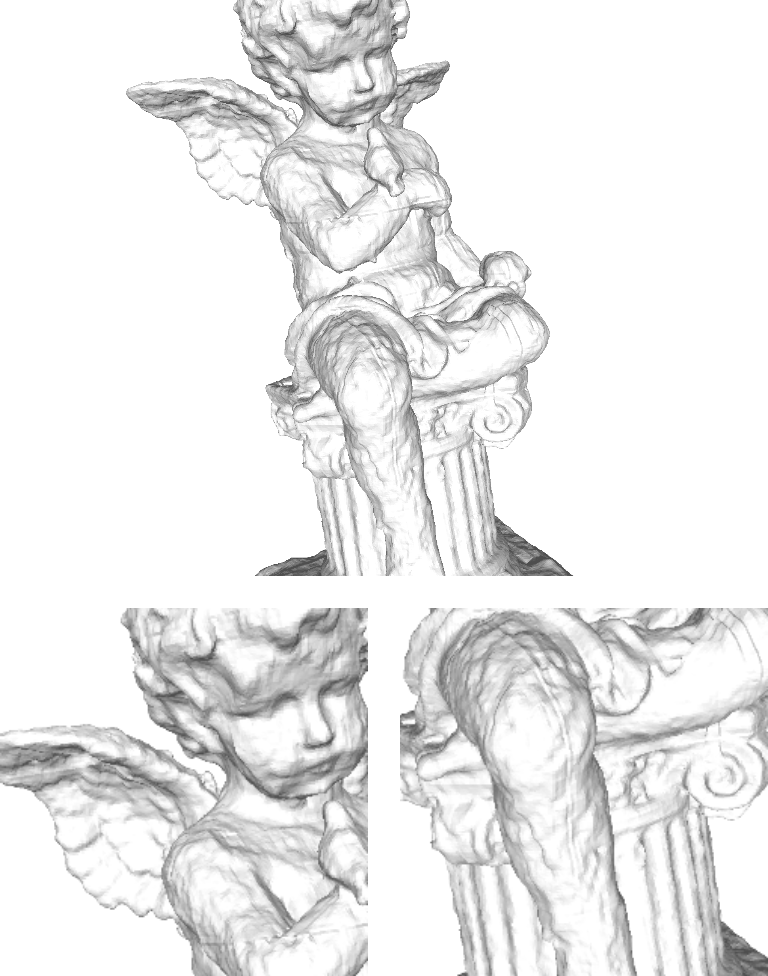}\vspace{2mm}\end{subfigure}
    \begin{subfigure}[b]{0.16\linewidth}\centering\includegraphics[width=\linewidth]{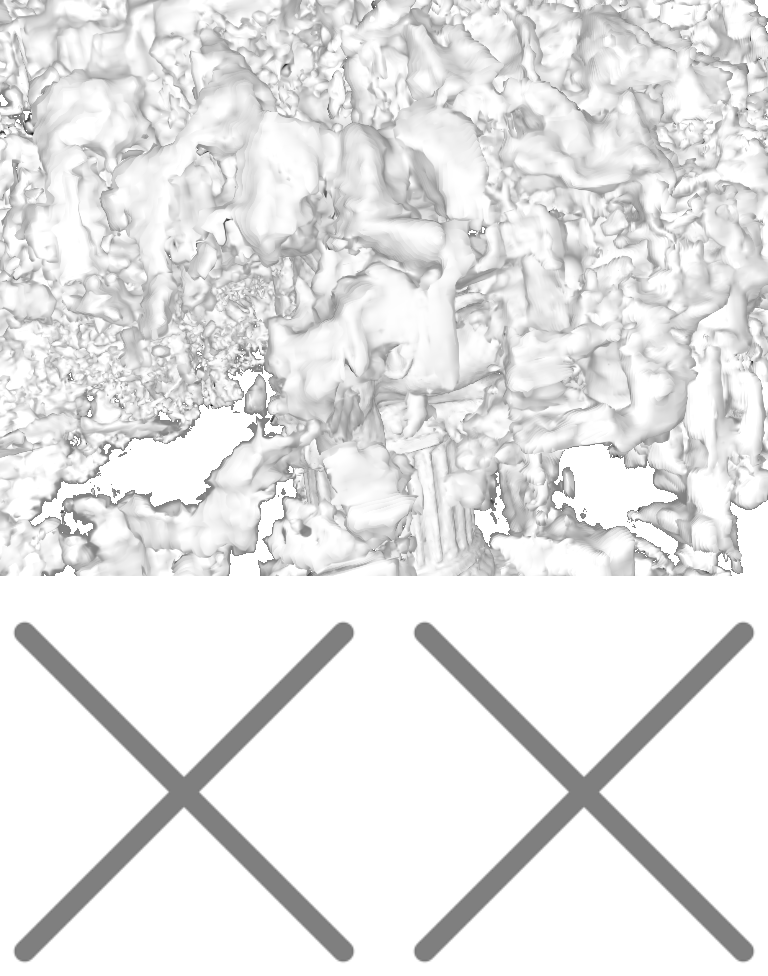}\vspace{2mm}\end{subfigure}
    \begin{subfigure}[b]{0.16\linewidth}\centering\includegraphics[width=\linewidth]{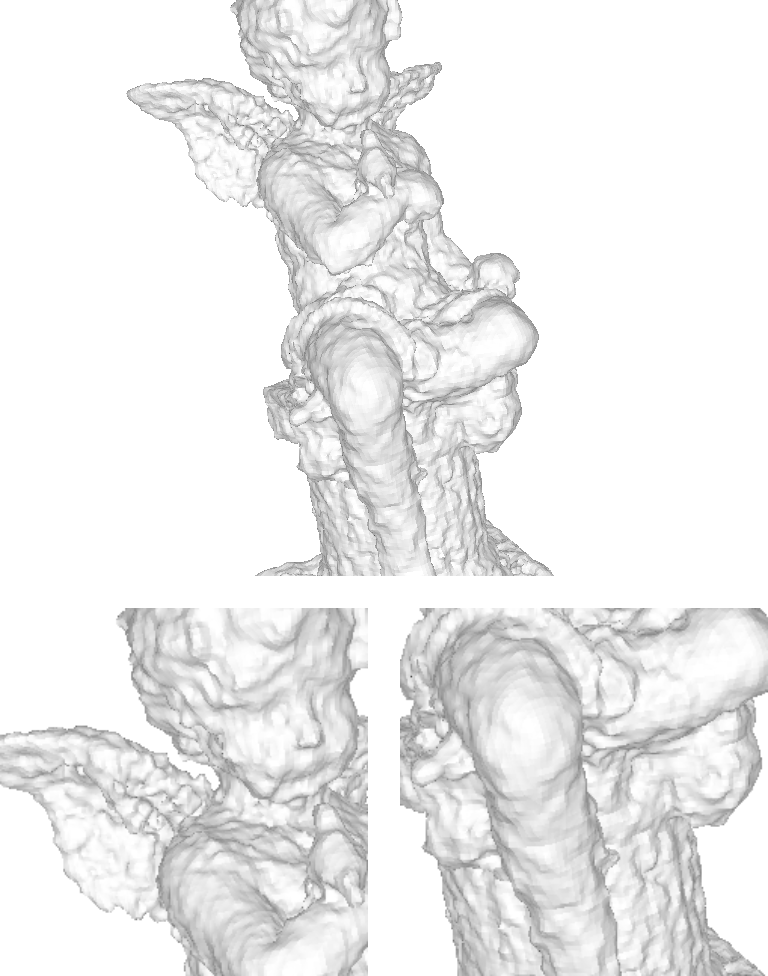}\vspace{2mm}\end{subfigure}
    \begin{subfigure}[b]{0.16\linewidth}\centering\includegraphics[width=\linewidth]{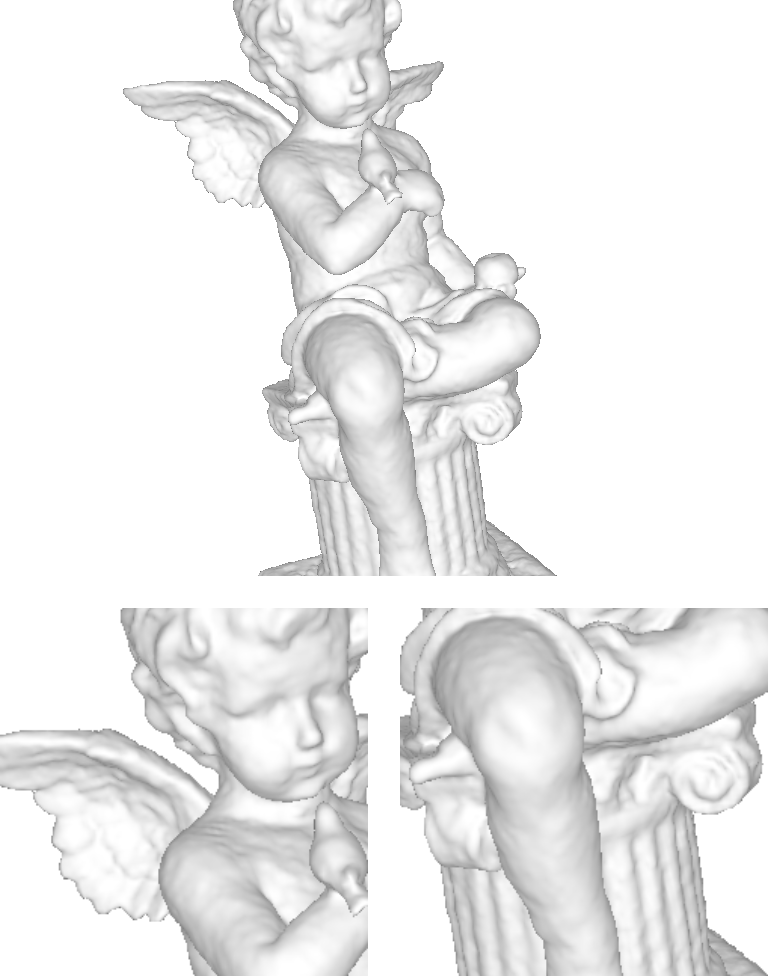}\vspace{2mm}\end{subfigure}

    \begin{subfigure}[b]{0.16\linewidth}\centering\includegraphics[width=\linewidth]{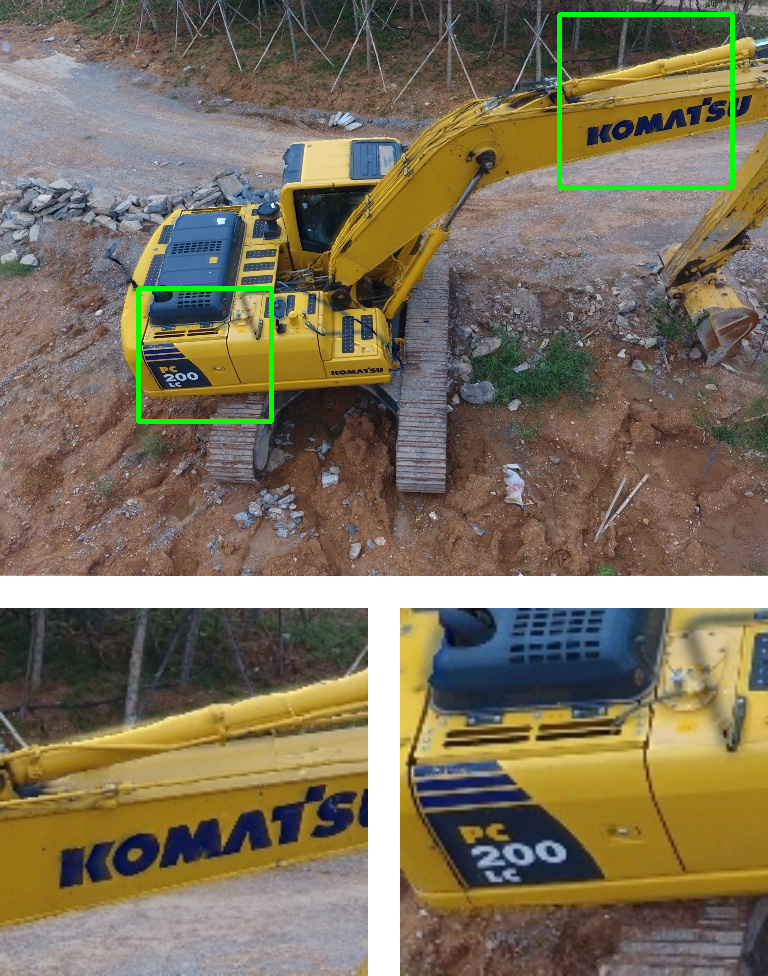}\caption{Ground Truth}\end{subfigure}
    \begin{subfigure}[b]{0.16\linewidth}\centering\includegraphics[width=\linewidth]{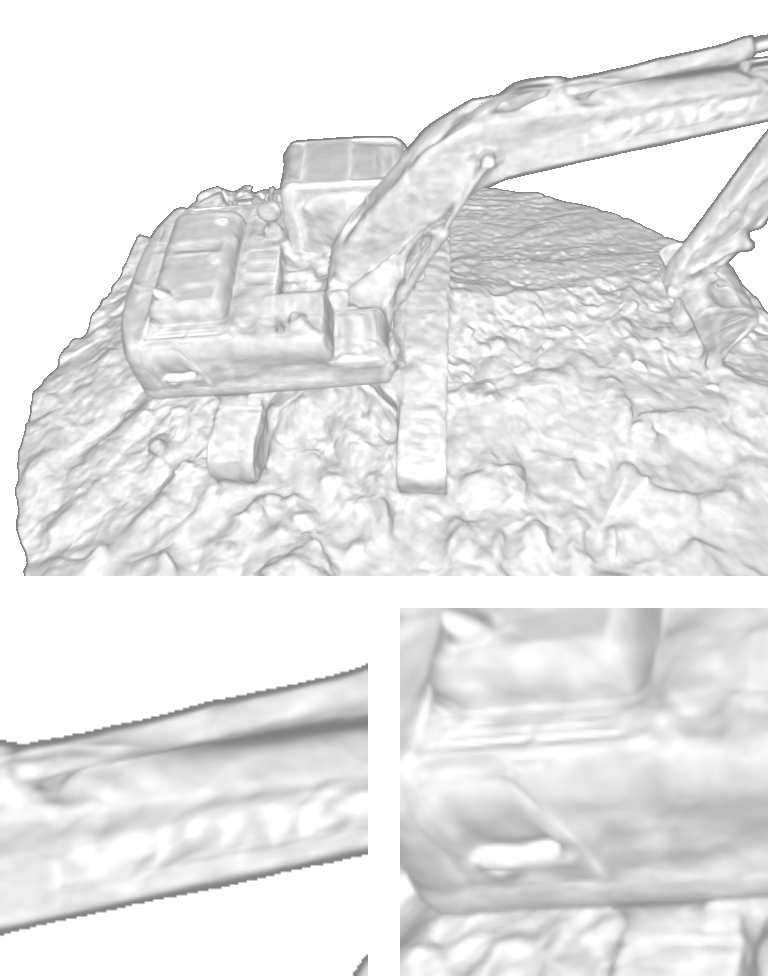}\caption{NeuS}\end{subfigure}
    \begin{subfigure}[b]{0.16\linewidth}\centering\includegraphics[width=\linewidth]{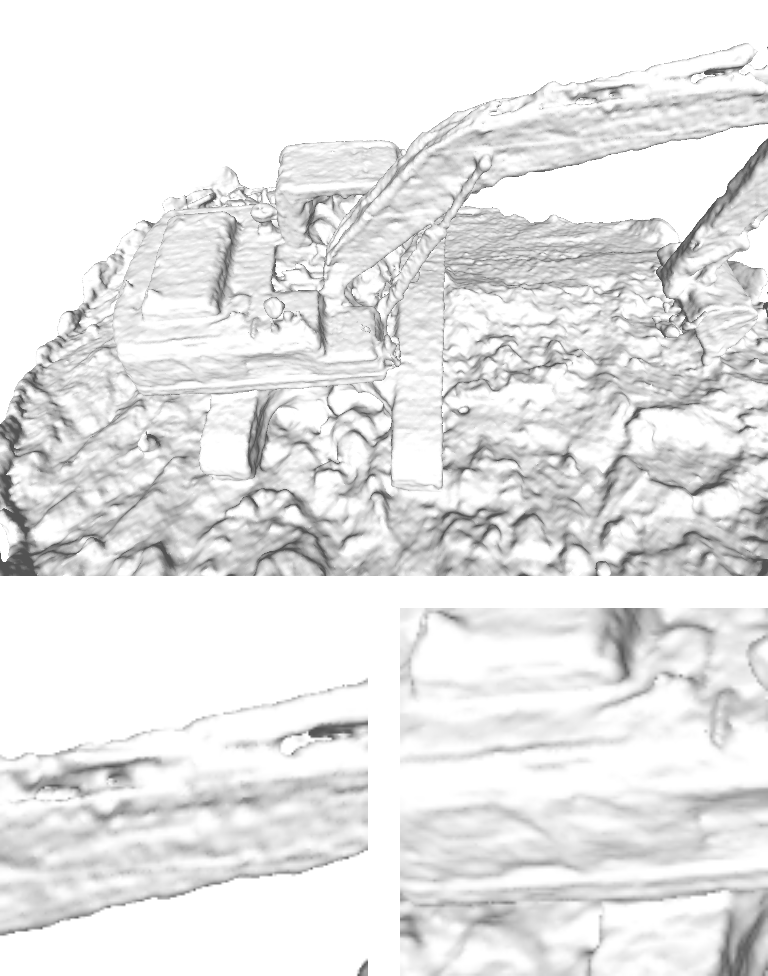}\caption{NeuS2}\end{subfigure}
    \begin{subfigure}[b]{0.16\linewidth}\centering\includegraphics[width=\linewidth]{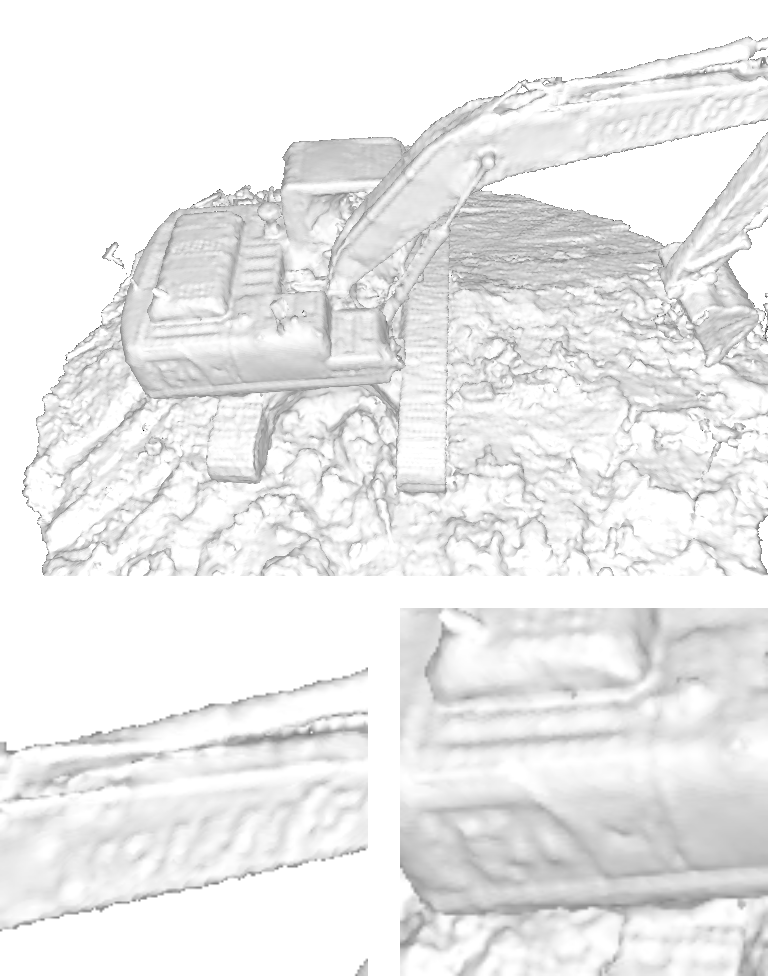}\caption{Voxurf}\end{subfigure}
    \begin{subfigure}[b]{0.16\linewidth}\centering\includegraphics[width=\linewidth]{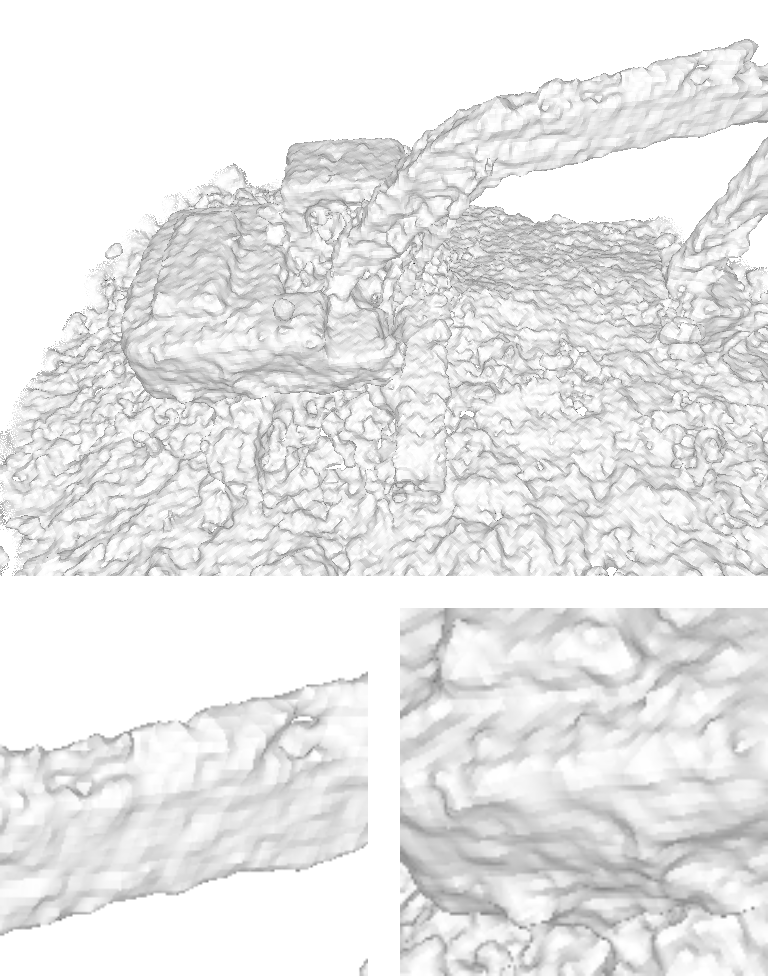}\caption{w/o Interp. Grad.}\end{subfigure}
    \begin{subfigure}[b]{0.16\linewidth}\centering\includegraphics[width=\linewidth]{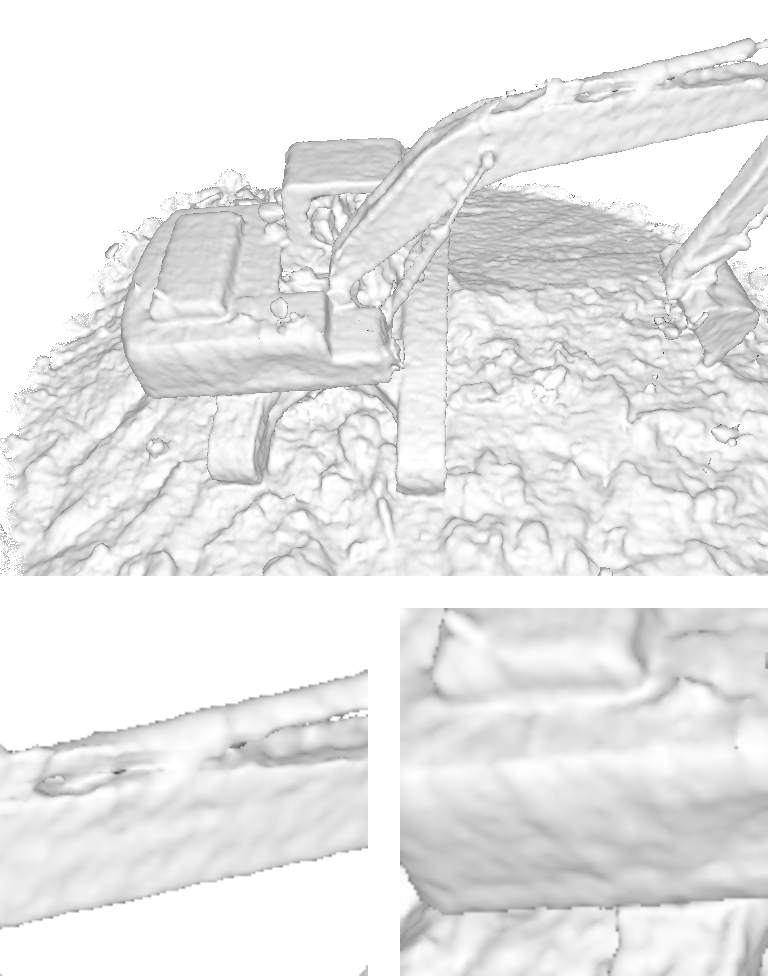}\caption{Ours}\end{subfigure}
    \caption{Visual comparison among different methods. All methods except NeuS2~\cite{wang2023neus2} are trained without foreground masks. The first two cases are from DTU~\cite{jensen2014large}, while the next two are from BlendedMVS~\cite{yao2020blendedmvs}. Voxurf failed to reconstruct the 3rd case due to the inability to distinguish foreground and foreground.}
    \label{fig:compare}
\end{figure*}

We report the quantitative results for surface reconstruction without foreground supervision on the DTU dataset on Tab.~\ref{tab:womask}. The results show that our method achieves a lower Chamfer Distance than previous methods under the same setting. We conduct a qualitative evaluation on DTU and BlendedMVS datasets respectively, without the supervision of foreground masks, as shown in Fig.~\ref{fig:compare}.
NeuS achieves a relatively smooth surface, but loses some high-frequency information, and the normal estimate with MLP derivatives is inaccurate. NeuS2 and Voxurf improve high-frequency details, but the surface is rough. All comparison methods are affected by surface texture when reconstructing planes, resulting in deformation. In comparison, our method obtains a smoother surface while retaining high-frequency details due to the gradient interpolation and geometry-radiance disentanglement.

We further compare the efficiency and performance with our main baselines Voxurf~\cite{wu2022voxurf} and NeuS2~\cite{wang2023neus2}, as shown in Tab.~\ref{tab:voxurf_neus2}. NeuS2 and our method are evaluated on 2080ti while Voxurf is evaluated on RTX 3090 to meet its memory consumption. Voxurf's two-stage training policy and complicated strategy make it computational and memory inefficient. NeuS2 implements the whole system in CUDA to avoid the computational overhead that the learning framework introduces, and directly calculates second-order derivatives for acceleration. Thus the 20k iterations of training cost 9 minutes and 6.2 GB of memory, but with the sacrifice of flexibility. Our method is mainly implemented in Python, and only key operators are accelerated using CUDA. Compared with NeuS2, the iterations of VoxNeuS are doubled, because the geometry-radiance disentangled architecture requires longer iterations to converge. Even so, VoxNeuS can still complete training in 15 minutes while using only 2.5 GB of memory. As for performance, VoxNeuS achieves the best novel-view PSNR and Chamfer Distance. NeuS2 is not applicable without foreground masks.

\subsection{Ablation Study}


\begin{table}[t]
    \caption{Ablation study on gradient interpolation.}
    \centering
    \begin{tabular}{c|cc}
        \toprule
        Gradient & Analytical & Interpolated \\ \midrule
        w/o mask & 0.93 & \textbf{0.72} \\
        w/ mask  & 0.82 & \textbf{0.70} \\
        \bottomrule
    \end{tabular}
    \label{tab:ig}
\end{table}

\begin{table*}[t]
\caption{Comparison among different methods without foreground mask supervision. The best and second best results are marked with \textbf{bold} and \underline{underline} respectively.}
\label{tab:womask}
\begin{tabular}{l|cccccccccccccccc}
\toprule
Scan   & 24   & 37   & 40   & 55   & 63   & 65   & 69   & 83   & 97   & 105  & 106  & 110  & 114  & 118  & 122  & mean \\
\midrule
NeRF\cite{mildenhall2021nerf}  & 1.90 & 1.60 & 1.85 & 0.58 & 2.28 & 1.27 & 1.47 & 1.67 & 2.05 & 1.07 & 0.88 & 2.53 & 1.06 & 1.15 & 0.96 & 1.49 \\
Colmap\cite{yu2022monosdf} & 0.81 & 2.05 & 0.73 & 1.22 & 1.79 & 1.58 & 1.02 & 3.05 & 1.40 & 2.05 & 1.00 & 1.32 & 0.49 & 0.78 & 1.17 & 1.36 \\
VolSDF\cite{yariv2021volume} & 1.14 & 1.26 & 0.81 & 0.49 & 1.25 & 0.70 & 0.72 & \underline{1.29} & 1.18 & \textbf{0.70} & 0.66 & \underline{1.08} & 0.42 & 0.61 & 0.55 & 0.86 \\
NeuS\cite{wang2021neus}   & 1.00 & 1.37 & 0.93 & 0.43 & 1.10 & \textbf{0.65} & \textbf{0.57} & 1.48 & \underline{1.09} & 0.83 & 0.52 & 1.20 & \textbf{0.35} & \textbf{0.49} & 0.54 & 0.84 \\
2DGS\cite{huang20242d} & \textbf{0.48} & 0.91 & \textbf{0.39} & \underline{0.39} & \textbf{1.01} & 0.83 & 0.81 & 1.36 & 1.27 & \underline{0.76} & 0.70 & 1.40 & \underline{0.40} & 0.76 & \underline{0.52} & 0.80 \\
Voxurf\cite{wu2022voxurf} & 0.72 & \textbf{0.75} & 0.47 & \underline{0.39} & 1.47 & \underline{0.76} & 0.81 & \textbf{1.02} & \textbf{1.04} & 0.92 & \underline{0.52} & 1.13 & \underline{0.40} & 0.53 & 0.53 & \underline{0.76} \\
\midrule
Ours   & \underline{0.58} & \underline{0.76} & \underline{0.40} & \textbf{0.36} & \underline{1.09} & \underline{0.76} & \underline{0.70} & \underline{1.29} & 1.15 & 0.84 & \textbf{0.44} & \textbf{1.02} & 0.42 & \underline{0.50} & \textbf{0.49} & \textbf{0.72} \\
\bottomrule
\end{tabular}
\end{table*}

\begin{table*}
    \caption{Comparison on efficiency and performance among Voxurf~\cite{wu2022voxurf}, NeuS2~\cite{wang2023neus2}, and our method.}
    \centering
    \begin{tabular}{c|cccc|ccc}
    \toprule
        Methods & Batches/s & Training Time & GPU Memory  & Implementation & PSNR & CD w/ mask & CD w/o mask\\ \midrule
        NeuS & 8.3 & 10 hours & 10 GB & Python & 29.63 & 0.77 & 0.84 \\
        Voxurf & 10.0 & 50 mins & 14 GB & Python \& CUDA/C++ & 32.16 & 0.72 & 0.76 \\
        NeuS2 & 37.0 & \textbf{9 mins} & 6.2 GB & CUDA/C++ & 32.14 & \textbf{0.70} & - \\
        Ours & \textbf{44.4} & 15 mins & \textbf{2.5 GB} & Python \& CUDA/C++ & \textbf{32.21} & \textbf{0.70} & \textbf{0.72} \\ \bottomrule
    \end{tabular}
    \label{tab:voxurf_neus2}
\end{table*}

This section evaluates our method's effectiveness in gradient interpolation and regularization terms.

We first evaluate the effectiveness of interpolated gradient on the DTU dataset with Chamfer Distance. The quantitative results are listed in Tab~\ref{tab:ig}. All hyper-parameters are kept the same except for the estimation strategy of the SDF gradient $\nabla_\myx f(\myx)$. Some qualitative comparison is shown in Fig.~\ref{fig:comp1} and Fig.~\ref{fig:compare}. Experimental results show that gradient interpolation is key to the correct reconstruction of our method. The glitches~(Fig.\ref{fig:example}) caused by the analytical gradient make the reconstructed surface discontinuous and rough.

\begin{figure}
    \centering
    \begin{subfigure}[b]{0.4\linewidth}
        \centering
        \includegraphics[width=\linewidth]{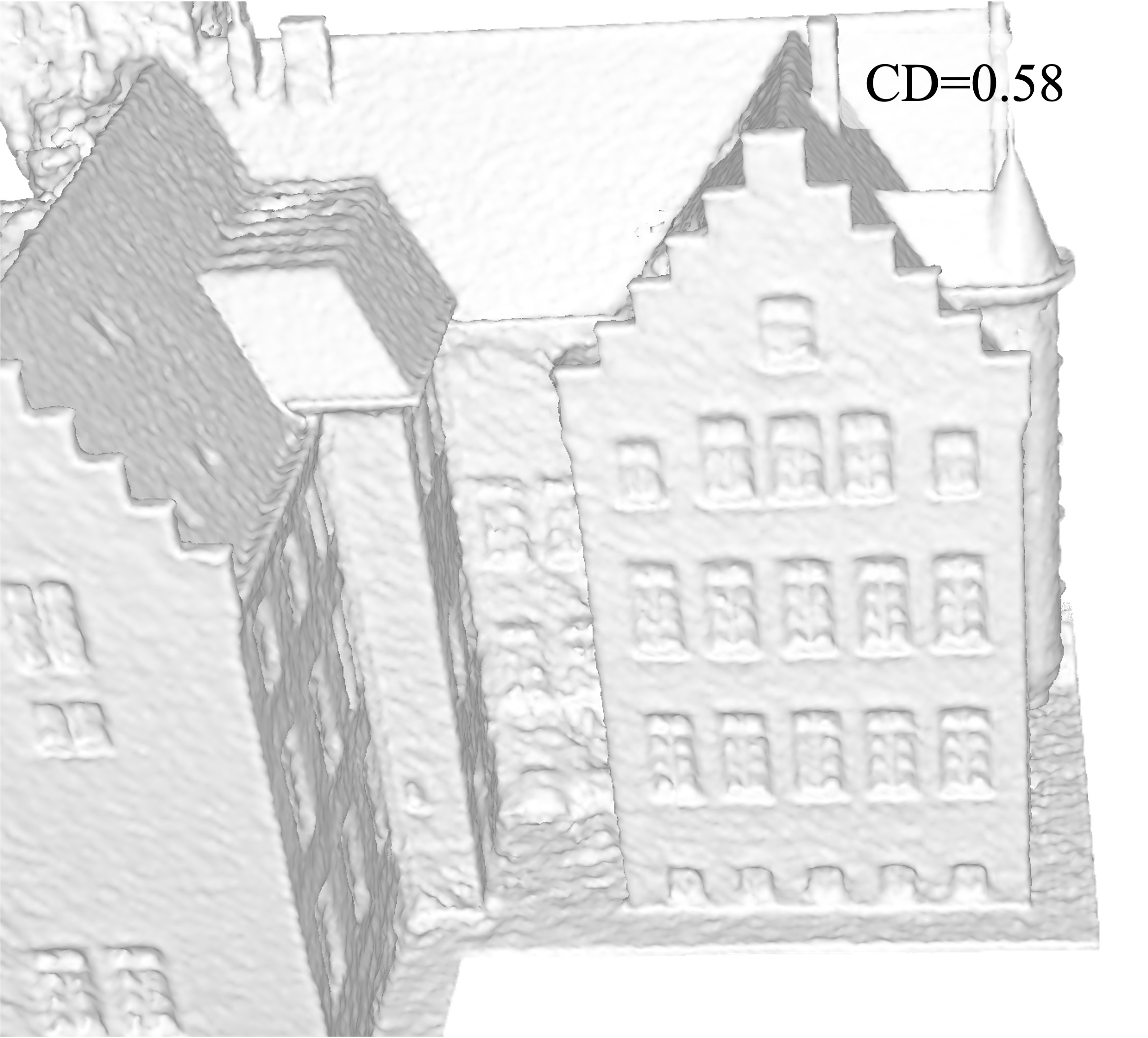}
        \caption{full}
    \end{subfigure}
    \begin{subfigure}[b]{0.4\linewidth}
        \centering
        \includegraphics[width=\linewidth]{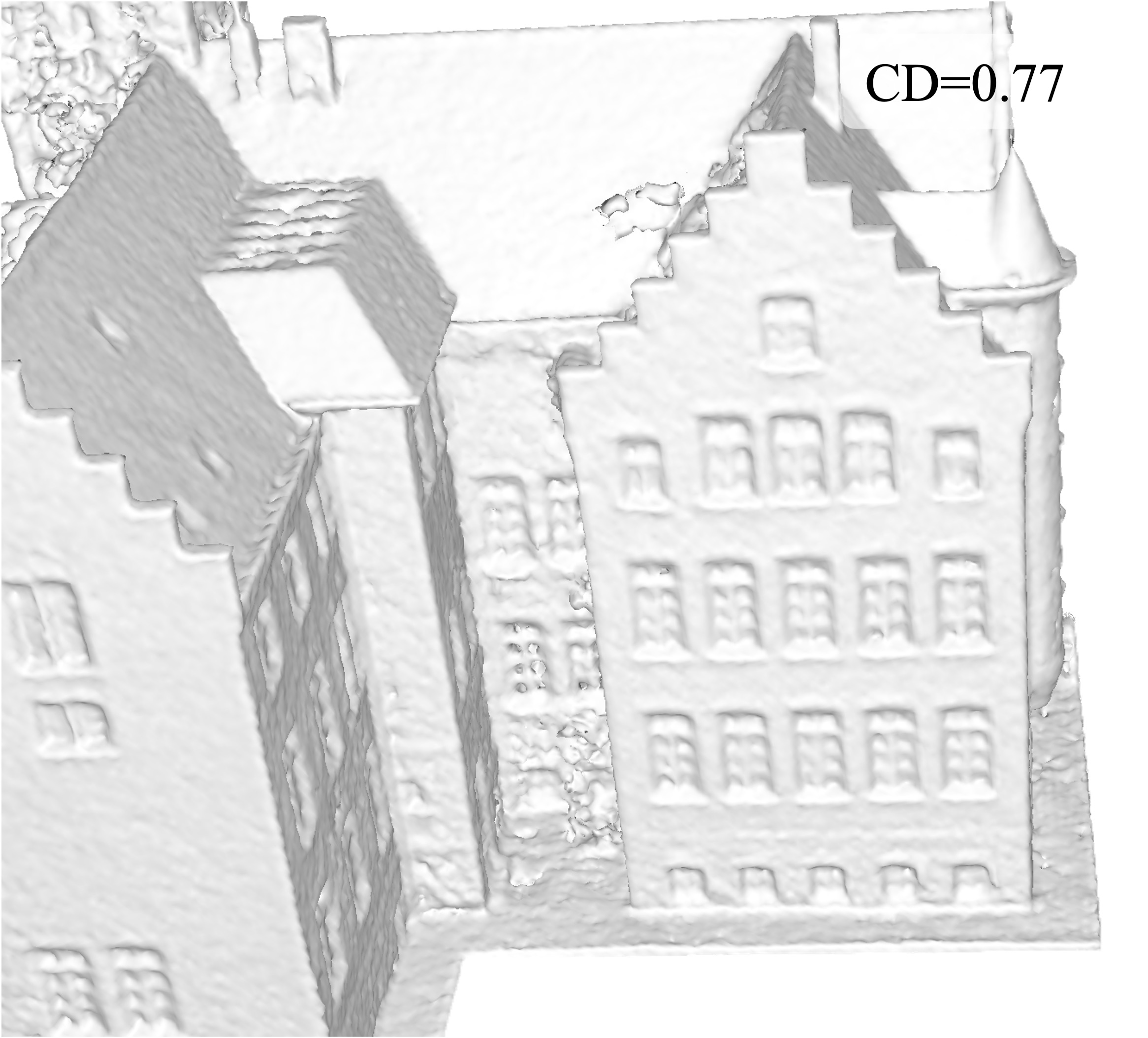}
        \caption{w/o curvature}
    \end{subfigure}
    \begin{subfigure}[b]{0.4\linewidth}
        \centering
        \includegraphics[width=\linewidth]{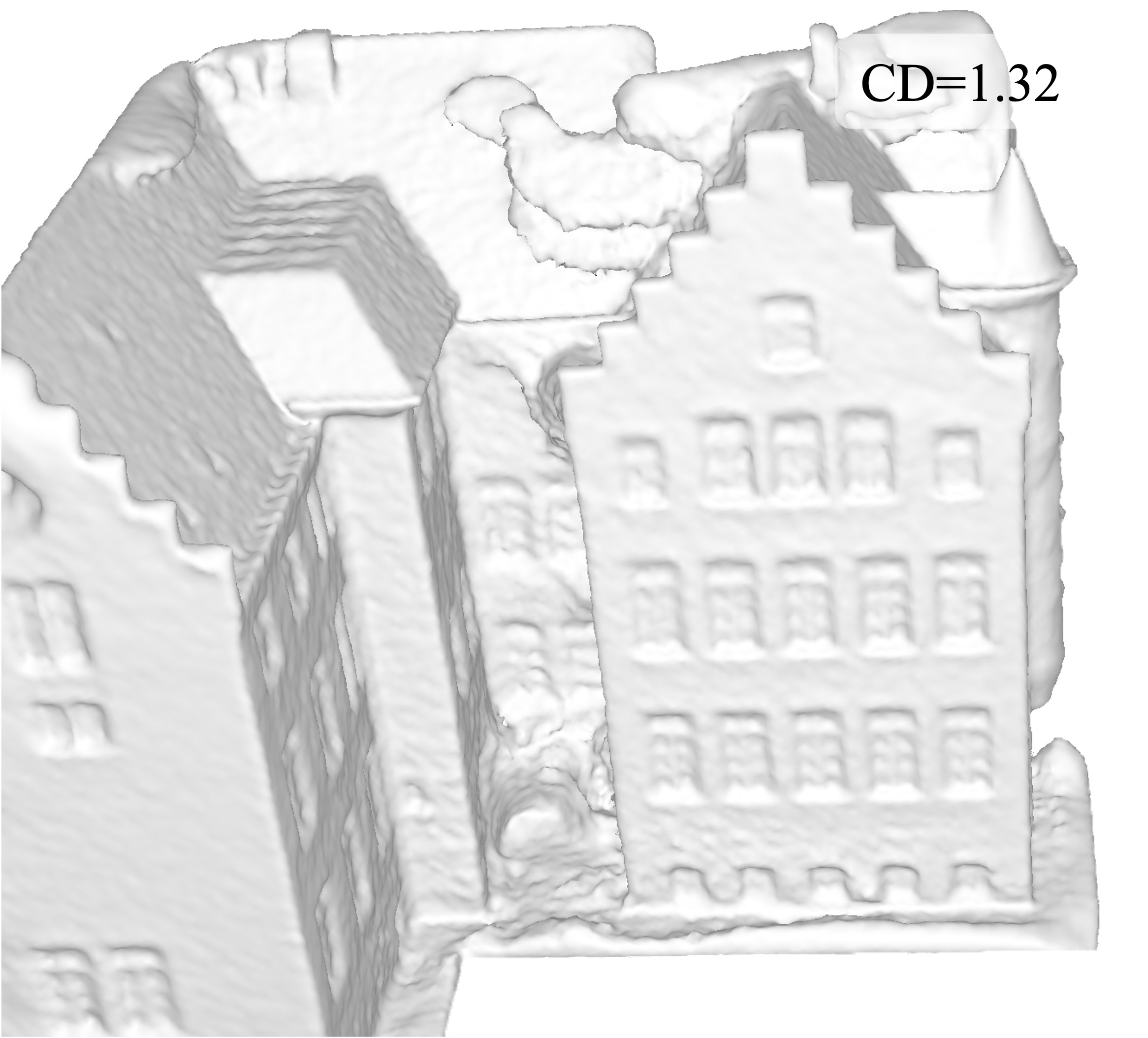}
        \caption{w/o Eikonal}
    \end{subfigure}
    \begin{subfigure}[b]{0.4\linewidth}
        \centering
        \includegraphics[width=\linewidth]{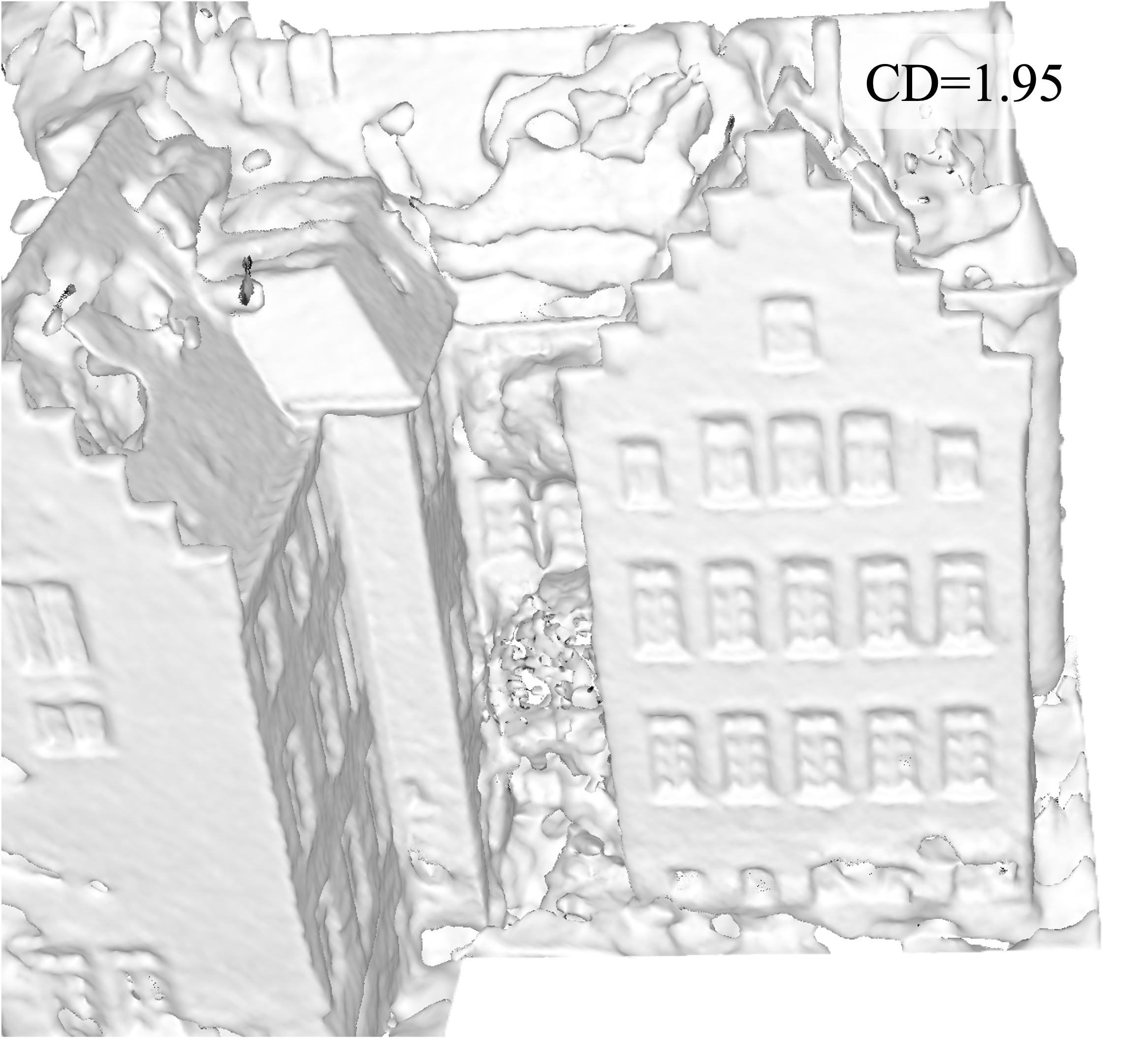}
        \caption{w/o both}
    \end{subfigure}
    \caption{Ablation study on regularization terms.}
    \label{fig:regularization}
\end{figure}

Then we qualitatively evaluate the effectiveness of regularization terms, taking a scene of the DTU dataset as an example. The reconstructions without Eikonal loss or/and curvature loss are illustrated in Fig~\ref{fig:regularization}. Previous MLP-based implicit representation provides strong low-frequency priors, making the reconstruction smooth. However, our voxel-based explicit representation makes the regularization terms particularly important for reconstruction. Curvature loss makes the surface smooth, while Eikonal loss makes the SDF grid converge normally.

\subsection{Efficiency Study}

\begin{figure}
    \centering
    \includegraphics[width=0.9\linewidth]{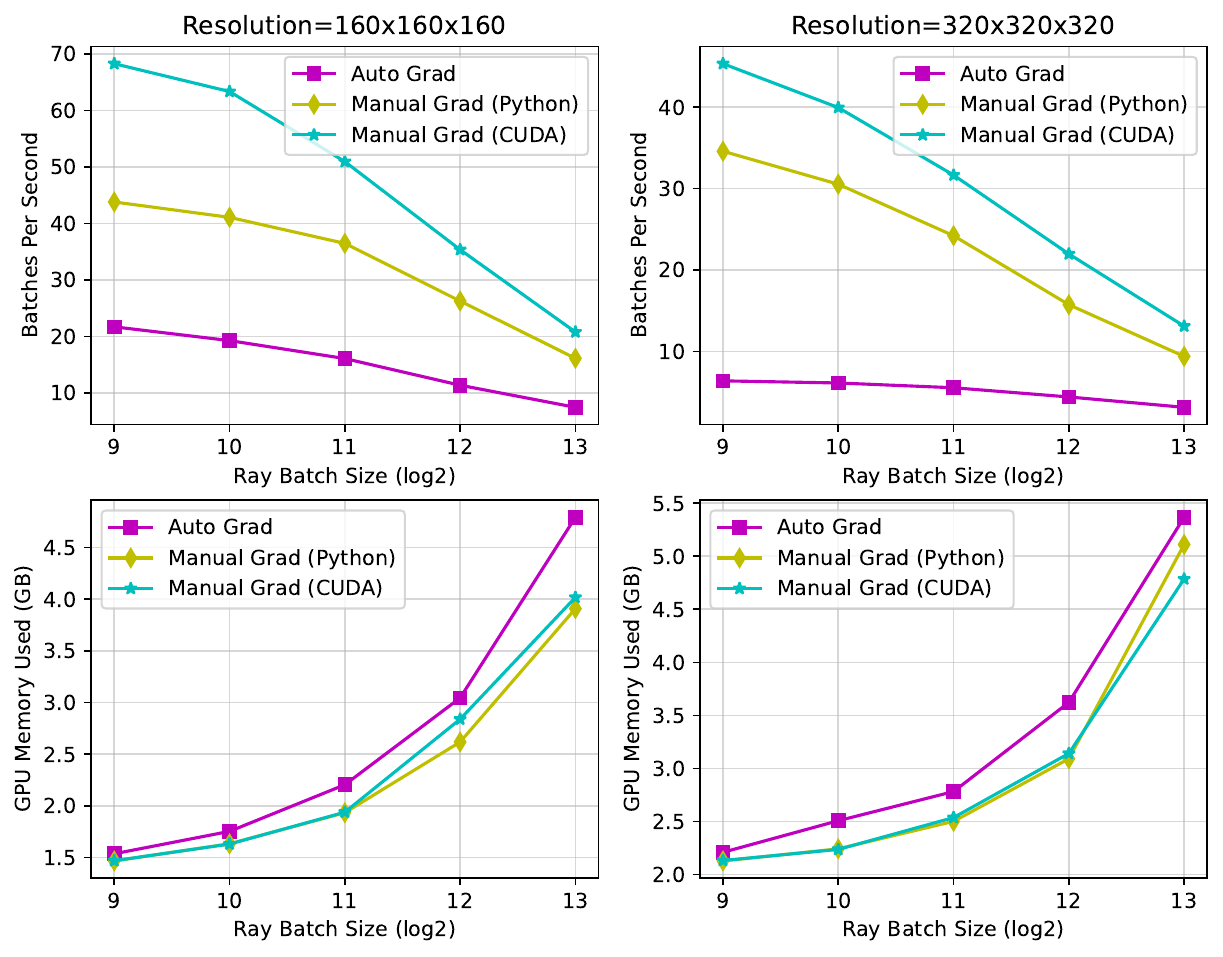}
    \caption{Comparison of speed and memory among different implementations for Eikonal and curvature regularization. Auto Grad: automatic differentiation provided by PyTorch; Manual Grad (Python): our explicit gradient solving in Python; Manual Grad (CUDA): ours in CUDA/C++. }
    \label{fig:timing}
\end{figure}

In this experiment, we evaluate the acceleration of our efficient SDF regularization, as shown in Fig.~\ref{fig:timing}. We compare the training speed and memory consumption under different regularization settings. When automatic differentiation is applied, regularization is the training bottleneck, taking up more than 70\% of the training time. This is because frequent unstructured memory access leads to cache misses, which in turn affects efficiency. Our explicit gradient solving reduces the frequency of memory accesses and computations, increasing the overall training speed by 2.2x. The CUDA implementation of explicit gradient solving eliminates the overhead caused by Python and improves the locality of memory access, thus bringing about 3x acceleration and eliminating this bottleneck. We set the ray batch size as 2048 to balance performance with time and memory overhead.

\section{Conclusion}

We propose VoxNeuS, an efficient neural surface reconstruction method, where SDF values are explicitly stored in a dense grid to improve efficiency and model high-frequency details. However, trilinear interpolation makes the gradient discontinuous, thereby degrading the reconstruction quality. Therefore, we propose interpolated gradients to eliminate discontinuities caused by original analytical gradients. To tackle the bottleneck caused by the automatic differentiation of regularization terms, we take advantage of explicit expression, explicitly solve its derivatives and implement it using CUDA, thus optimizing the memory access frequency and locality. In terms of network architecture, we found that using the same network to fit geometry and radiance, or entangling the geometry and radiance in the computation graph, makes the geometric structure deform to fit the radiance. Therefore, we designed a geometry-radiance disentangled architecture to eliminate this artifact.
The experiments illustrate the efficiency and effectiveness of our method.


\bibliographystyle{ACM-Reference-Format}
\bibliography{sample-base}


\begin{thebibliography}{34}


\ifx \showCODEN    \undefined \def \showCODEN     #1{\unskip}     \fi
\ifx \showDOI      \undefined \def \showDOI       #1{#1}\fi
\ifx \showISBNx    \undefined \def \showISBNx     #1{\unskip}     \fi
\ifx \showISBNxiii \undefined \def \showISBNxiii  #1{\unskip}     \fi
\ifx \showISSN     \undefined \def \showISSN      #1{\unskip}     \fi
\ifx \showLCCN     \undefined \def \showLCCN      #1{\unskip}     \fi
\ifx \shownote     \undefined \def \shownote      #1{#1}          \fi
\ifx \showarticletitle \undefined \def \showarticletitle #1{#1}   \fi
\ifx \showURL      \undefined \def \showURL       {\relax}        \fi
\providecommand\bibfield[2]{#2}
\providecommand\bibinfo[2]{#2}
\providecommand\natexlab[1]{#1}
\providecommand\showeprint[2][]{arXiv:#2}

\bibitem[Barrett and Dherin(2020)]%
        {barrett2020implicit}
\bibfield{author}{\bibinfo{person}{David~GT Barrett} {and}
  \bibinfo{person}{Benoit Dherin}.} \bibinfo{year}{2020}\natexlab{}.
\newblock \showarticletitle{Implicit gradient regularization}.
\newblock \bibinfo{journal}{\emph{arXiv preprint arXiv:2009.11162}}
  (\bibinfo{year}{2020}).
\newblock


\bibitem[Cai et~al\mbox{.}(2023)]%
        {cai2023neuda}
\bibfield{author}{\bibinfo{person}{Bowen Cai}, \bibinfo{person}{Jinchi Huang},
  \bibinfo{person}{Rongfei Jia}, \bibinfo{person}{Chengfei Lv}, {and}
  \bibinfo{person}{Huan Fu}.} \bibinfo{year}{2023}\natexlab{}.
\newblock \showarticletitle{Neuda: Neural deformable anchor for high-fidelity
  implicit surface reconstruction}. In \bibinfo{booktitle}{\emph{Proceedings of
  the IEEE/CVF Conference on Computer Vision and Pattern Recognition}}.
  \bibinfo{pages}{8476--8485}.
\newblock


\bibitem[Darmon et~al\mbox{.}(2022)]%
        {darmon2022improving}
\bibfield{author}{\bibinfo{person}{Fran{\c{c}}ois Darmon},
  \bibinfo{person}{B{\'e}n{\'e}dicte Bascle}, \bibinfo{person}{Jean-Cl{\'e}ment
  Devaux}, \bibinfo{person}{Pascal Monasse}, {and} \bibinfo{person}{Mathieu
  Aubry}.} \bibinfo{year}{2022}\natexlab{}.
\newblock \showarticletitle{Improving neural implicit surfaces geometry with
  patch warping}. In \bibinfo{booktitle}{\emph{Proceedings of the IEEE/CVF
  Conference on Computer Vision and Pattern Recognition}}.
  \bibinfo{pages}{6260--6269}.
\newblock


\bibitem[Fridovich-Keil et~al\mbox{.}(2022)]%
        {fridovich2022plenoxels}
\bibfield{author}{\bibinfo{person}{Sara Fridovich-Keil}, \bibinfo{person}{Alex
  Yu}, \bibinfo{person}{Matthew Tancik}, \bibinfo{person}{Qinhong Chen},
  \bibinfo{person}{Benjamin Recht}, {and} \bibinfo{person}{Angjoo Kanazawa}.}
  \bibinfo{year}{2022}\natexlab{}.
\newblock \showarticletitle{Plenoxels: Radiance fields without neural
  networks}. In \bibinfo{booktitle}{\emph{Proceedings of the IEEE/CVF
  Conference on Computer Vision and Pattern Recognition}}.
  \bibinfo{pages}{5501--5510}.
\newblock


\bibitem[Green(2007)]%
        {green2007improved}
\bibfield{author}{\bibinfo{person}{Chris Green}.}
  \bibinfo{year}{2007}\natexlab{}.
\newblock \showarticletitle{Improved alpha-tested magnification for vector
  textures and special effects}.
\newblock In \bibinfo{booktitle}{\emph{ACM SIGGRAPH 2007 courses}}.
  \bibinfo{pages}{9--18}.
\newblock


\bibitem[Gropp et~al\mbox{.}(2020)]%
        {gropp2020implicit}
\bibfield{author}{\bibinfo{person}{Amos Gropp}, \bibinfo{person}{Lior Yariv},
  \bibinfo{person}{Niv Haim}, \bibinfo{person}{Matan Atzmon}, {and}
  \bibinfo{person}{Yaron Lipman}.} \bibinfo{year}{2020}\natexlab{}.
\newblock \showarticletitle{Implicit geometric regularization for learning
  shapes}.
\newblock \bibinfo{journal}{\emph{arXiv preprint arXiv:2002.10099}}
  (\bibinfo{year}{2020}).
\newblock


\bibitem[Gu{\'e}don and Lepetit(2023)]%
        {guedon2023sugar}
\bibfield{author}{\bibinfo{person}{Antoine Gu{\'e}don} {and}
  \bibinfo{person}{Vincent Lepetit}.} \bibinfo{year}{2023}\natexlab{}.
\newblock \showarticletitle{Sugar: Surface-aligned gaussian splatting for
  efficient 3d mesh reconstruction and high-quality mesh rendering}.
\newblock \bibinfo{journal}{\emph{arXiv preprint arXiv:2311.12775}}
  (\bibinfo{year}{2023}).
\newblock


\bibitem[Hartley and Zisserman(2003)]%
        {hartley2003multiple}
\bibfield{author}{\bibinfo{person}{Richard Hartley} {and}
  \bibinfo{person}{Andrew Zisserman}.} \bibinfo{year}{2003}\natexlab{}.
\newblock \bibinfo{booktitle}{\emph{Multiple view geometry in computer
  vision}}.
\newblock \bibinfo{publisher}{Cambridge university press}.
\newblock


\bibitem[Huang et~al\mbox{.}(2024)]%
        {huang20242d}
\bibfield{author}{\bibinfo{person}{Binbin Huang}, \bibinfo{person}{Zehao Yu},
  \bibinfo{person}{Anpei Chen}, \bibinfo{person}{Andreas Geiger}, {and}
  \bibinfo{person}{Shenghua Gao}.} \bibinfo{year}{2024}\natexlab{}.
\newblock \showarticletitle{2D Gaussian Splatting for Geometrically Accurate
  Radiance Fields}.
\newblock \bibinfo{journal}{\emph{arXiv preprint arXiv:2403.17888}}
  (\bibinfo{year}{2024}).
\newblock


\bibitem[Jensen et~al\mbox{.}(2014)]%
        {jensen2014large}
\bibfield{author}{\bibinfo{person}{Rasmus Jensen}, \bibinfo{person}{Anders
  Dahl}, \bibinfo{person}{George Vogiatzis}, \bibinfo{person}{Engin Tola},
  {and} \bibinfo{person}{Henrik Aan{\ae}s}.} \bibinfo{year}{2014}\natexlab{}.
\newblock \showarticletitle{Large scale multi-view stereopsis evaluation}. In
  \bibinfo{booktitle}{\emph{Proceedings of the IEEE conference on computer
  vision and pattern recognition}}. \bibinfo{pages}{406--413}.
\newblock


\bibitem[Kerbl et~al\mbox{.}(2023)]%
        {kerbl20233d}
\bibfield{author}{\bibinfo{person}{Bernhard Kerbl}, \bibinfo{person}{Georgios
  Kopanas}, \bibinfo{person}{Thomas Leimk{\"u}hler}, {and}
  \bibinfo{person}{George Drettakis}.} \bibinfo{year}{2023}\natexlab{}.
\newblock \showarticletitle{3d gaussian splatting for real-time radiance field
  rendering}.
\newblock \bibinfo{journal}{\emph{ACM Transactions on Graphics}}
  \bibinfo{volume}{42}, \bibinfo{number}{4} (\bibinfo{year}{2023}),
  \bibinfo{pages}{1--14}.
\newblock


\bibitem[Li et~al\mbox{.}(2022)]%
        {li2022vox}
\bibfield{author}{\bibinfo{person}{Hai Li}, \bibinfo{person}{Xingrui Yang},
  \bibinfo{person}{Hongjia Zhai}, \bibinfo{person}{Yuqian Liu},
  \bibinfo{person}{Hujun Bao}, {and} \bibinfo{person}{Guofeng Zhang}.}
  \bibinfo{year}{2022}\natexlab{}.
\newblock \showarticletitle{Vox-surf: Voxel-based implicit surface
  representation}.
\newblock \bibinfo{journal}{\emph{IEEE Transactions on Visualization and
  Computer Graphics}} (\bibinfo{year}{2022}).
\newblock


\bibitem[Li et~al\mbox{.}(2023)]%
        {li2023neuralangelo}
\bibfield{author}{\bibinfo{person}{Zhaoshuo Li}, \bibinfo{person}{Thomas
  M{\"u}ller}, \bibinfo{person}{Alex Evans}, \bibinfo{person}{Russell~H
  Taylor}, \bibinfo{person}{Mathias Unberath}, \bibinfo{person}{Ming-Yu Liu},
  {and} \bibinfo{person}{Chen-Hsuan Lin}.} \bibinfo{year}{2023}\natexlab{}.
\newblock \showarticletitle{Neuralangelo: High-Fidelity Neural Surface
  Reconstruction}. In \bibinfo{booktitle}{\emph{Proceedings of the IEEE/CVF
  Conference on Computer Vision and Pattern Recognition}}.
  \bibinfo{pages}{8456--8465}.
\newblock


\bibitem[Mildenhall et~al\mbox{.}(2021)]%
        {mildenhall2021nerf}
\bibfield{author}{\bibinfo{person}{Ben Mildenhall}, \bibinfo{person}{Pratul~P
  Srinivasan}, \bibinfo{person}{Matthew Tancik}, \bibinfo{person}{Jonathan~T
  Barron}, \bibinfo{person}{Ravi Ramamoorthi}, {and} \bibinfo{person}{Ren Ng}.}
  \bibinfo{year}{2021}\natexlab{}.
\newblock \showarticletitle{Nerf: Representing scenes as neural radiance fields
  for view synthesis}.
\newblock \bibinfo{journal}{\emph{Commun. ACM}} \bibinfo{volume}{65},
  \bibinfo{number}{1} (\bibinfo{year}{2021}), \bibinfo{pages}{99--106}.
\newblock


\bibitem[M{\"u}ller et~al\mbox{.}(2022)]%
        {muller2022instant}
\bibfield{author}{\bibinfo{person}{Thomas M{\"u}ller}, \bibinfo{person}{Alex
  Evans}, \bibinfo{person}{Christoph Schied}, {and} \bibinfo{person}{Alexander
  Keller}.} \bibinfo{year}{2022}\natexlab{}.
\newblock \showarticletitle{Instant neural graphics primitives with a
  multiresolution hash encoding}.
\newblock \bibinfo{journal}{\emph{ACM Transactions on Graphics (ToG)}}
  \bibinfo{volume}{41}, \bibinfo{number}{4} (\bibinfo{year}{2022}),
  \bibinfo{pages}{1--15}.
\newblock


\bibitem[Niemeyer et~al\mbox{.}(2020)]%
        {niemeyer2020differentiable}
\bibfield{author}{\bibinfo{person}{Michael Niemeyer}, \bibinfo{person}{Lars
  Mescheder}, \bibinfo{person}{Michael Oechsle}, {and} \bibinfo{person}{Andreas
  Geiger}.} \bibinfo{year}{2020}\natexlab{}.
\newblock \showarticletitle{Differentiable volumetric rendering: Learning
  implicit 3d representations without 3d supervision}. In
  \bibinfo{booktitle}{\emph{Proceedings of the IEEE/CVF Conference on Computer
  Vision and Pattern Recognition}}. \bibinfo{pages}{3504--3515}.
\newblock


\bibitem[Rosu and Behnke(2023)]%
        {rosu2023permutosdf}
\bibfield{author}{\bibinfo{person}{Radu~Alexandru Rosu} {and}
  \bibinfo{person}{Sven Behnke}.} \bibinfo{year}{2023}\natexlab{}.
\newblock \showarticletitle{Permutosdf: Fast multi-view reconstruction with
  implicit surfaces using permutohedral lattices}. In
  \bibinfo{booktitle}{\emph{Proceedings of the IEEE/CVF Conference on Computer
  Vision and Pattern Recognition}}. \bibinfo{pages}{8466--8475}.
\newblock


\bibitem[Sch\"{o}nberger and Frahm(2016)]%
        {schoenberger2016sfm}
\bibfield{author}{\bibinfo{person}{Johannes~Lutz Sch\"{o}nberger} {and}
  \bibinfo{person}{Jan-Michael Frahm}.} \bibinfo{year}{2016}\natexlab{}.
\newblock \showarticletitle{Structure-from-Motion Revisited}. In
  \bibinfo{booktitle}{\emph{Conference on Computer Vision and Pattern
  Recognition (CVPR)}}.
\newblock


\bibitem[Sun et~al\mbox{.}(2022)]%
        {sun2022direct}
\bibfield{author}{\bibinfo{person}{Cheng Sun}, \bibinfo{person}{Min Sun}, {and}
  \bibinfo{person}{Hwann-Tzong Chen}.} \bibinfo{year}{2022}\natexlab{}.
\newblock \showarticletitle{Direct voxel grid optimization: Super-fast
  convergence for radiance fields reconstruction}. In
  \bibinfo{booktitle}{\emph{Proceedings of the IEEE/CVF Conference on Computer
  Vision and Pattern Recognition}}. \bibinfo{pages}{5459--5469}.
\newblock


\bibitem[Tian et~al\mbox{.}(2023)]%
        {tian2023superudf}
\bibfield{author}{\bibinfo{person}{Hui Tian}, \bibinfo{person}{Chenyang Zhu},
  \bibinfo{person}{Yifei Shi}, {and} \bibinfo{person}{Kai Xu}.}
  \bibinfo{year}{2023}\natexlab{}.
\newblock \showarticletitle{Superudf: Self-supervised udf estimation for
  surface reconstruction}.
\newblock \bibinfo{journal}{\emph{IEEE Transactions on Visualization and
  Computer Graphics}} (\bibinfo{year}{2023}).
\newblock


\bibitem[Wang et~al\mbox{.}(2021)]%
        {wang2021neus}
\bibfield{author}{\bibinfo{person}{Peng Wang}, \bibinfo{person}{Lingjie Liu},
  \bibinfo{person}{Yuan Liu}, \bibinfo{person}{Christian Theobalt},
  \bibinfo{person}{Taku Komura}, {and} \bibinfo{person}{Wenping Wang}.}
  \bibinfo{year}{2021}\natexlab{}.
\newblock \showarticletitle{Neus: Learning neural implicit surfaces by volume
  rendering for multi-view reconstruction}.
\newblock \bibinfo{journal}{\emph{arXiv preprint arXiv:2106.10689}}
  (\bibinfo{year}{2021}).
\newblock


\bibitem[Wang et~al\mbox{.}(2023a)]%
        {wang2023neus2}
\bibfield{author}{\bibinfo{person}{Yiming Wang}, \bibinfo{person}{Qin Han},
  \bibinfo{person}{Marc Habermann}, \bibinfo{person}{Kostas Daniilidis},
  \bibinfo{person}{Christian Theobalt}, {and} \bibinfo{person}{Lingjie Liu}.}
  \bibinfo{year}{2023}\natexlab{a}.
\newblock \showarticletitle{Neus2: Fast learning of neural implicit surfaces
  for multi-view reconstruction}. In \bibinfo{booktitle}{\emph{Proceedings of
  the IEEE/CVF International Conference on Computer Vision}}.
  \bibinfo{pages}{3295--3306}.
\newblock


\bibitem[Wang et~al\mbox{.}(2022)]%
        {wang2022hf}
\bibfield{author}{\bibinfo{person}{Yiqun Wang}, \bibinfo{person}{Ivan
  Skorokhodov}, {and} \bibinfo{person}{Peter Wonka}.}
  \bibinfo{year}{2022}\natexlab{}.
\newblock \showarticletitle{Hf-neus: Improved surface reconstruction using
  high-frequency details}.
\newblock \bibinfo{journal}{\emph{Advances in Neural Information Processing
  Systems}}  \bibinfo{volume}{35} (\bibinfo{year}{2022}),
  \bibinfo{pages}{1966--1978}.
\newblock


\bibitem[Wang et~al\mbox{.}(2023b)]%
        {wang2023pet}
\bibfield{author}{\bibinfo{person}{Yiqun Wang}, \bibinfo{person}{Ivan
  Skorokhodov}, {and} \bibinfo{person}{Peter Wonka}.}
  \bibinfo{year}{2023}\natexlab{b}.
\newblock \showarticletitle{Pet-neus: Positional encoding tri-planes for neural
  surfaces}. In \bibinfo{booktitle}{\emph{Proceedings of the IEEE/CVF
  Conference on Computer Vision and Pattern Recognition}}.
  \bibinfo{pages}{12598--12607}.
\newblock


\bibitem[Wu et~al\mbox{.}(2022)]%
        {wu2022voxurf}
\bibfield{author}{\bibinfo{person}{Tong Wu}, \bibinfo{person}{Jiaqi Wang},
  \bibinfo{person}{Xingang Pan}, \bibinfo{person}{Xudong Xu},
  \bibinfo{person}{Christian Theobalt}, \bibinfo{person}{Ziwei Liu}, {and}
  \bibinfo{person}{Dahua Lin}.} \bibinfo{year}{2022}\natexlab{}.
\newblock \showarticletitle{Voxurf: Voxel-based efficient and accurate neural
  surface reconstruction}.
\newblock \bibinfo{journal}{\emph{arXiv preprint arXiv:2208.12697}}
  (\bibinfo{year}{2022}).
\newblock


\bibitem[Yao et~al\mbox{.}(2020)]%
        {yao2020blendedmvs}
\bibfield{author}{\bibinfo{person}{Yao Yao}, \bibinfo{person}{Zixin Luo},
  \bibinfo{person}{Shiwei Li}, \bibinfo{person}{Jingyang Zhang},
  \bibinfo{person}{Yufan Ren}, \bibinfo{person}{Lei Zhou},
  \bibinfo{person}{Tian Fang}, {and} \bibinfo{person}{Long Quan}.}
  \bibinfo{year}{2020}\natexlab{}.
\newblock \showarticletitle{Blendedmvs: A large-scale dataset for generalized
  multi-view stereo networks}. In \bibinfo{booktitle}{\emph{Proceedings of the
  IEEE/CVF conference on computer vision and pattern recognition}}.
  \bibinfo{pages}{1790--1799}.
\newblock


\bibitem[Yariv et~al\mbox{.}(2021)]%
        {yariv2021volume}
\bibfield{author}{\bibinfo{person}{Lior Yariv}, \bibinfo{person}{Jiatao Gu},
  \bibinfo{person}{Yoni Kasten}, {and} \bibinfo{person}{Yaron Lipman}.}
  \bibinfo{year}{2021}\natexlab{}.
\newblock \showarticletitle{Volume rendering of neural implicit surfaces}.
\newblock \bibinfo{journal}{\emph{Advances in Neural Information Processing
  Systems}}  \bibinfo{volume}{34} (\bibinfo{year}{2021}),
  \bibinfo{pages}{4805--4815}.
\newblock


\bibitem[Yariv et~al\mbox{.}(2023)]%
        {yariv2023bakedsdf}
\bibfield{author}{\bibinfo{person}{Lior Yariv}, \bibinfo{person}{Peter Hedman},
  \bibinfo{person}{Christian Reiser}, \bibinfo{person}{Dor Verbin},
  \bibinfo{person}{Pratul~P Srinivasan}, \bibinfo{person}{Richard Szeliski},
  \bibinfo{person}{Jonathan~T Barron}, {and} \bibinfo{person}{Ben Mildenhall}.}
  \bibinfo{year}{2023}\natexlab{}.
\newblock \showarticletitle{BakedSDF: Meshing Neural SDFs for Real-Time View
  Synthesis}.
\newblock \bibinfo{journal}{\emph{arXiv preprint arXiv:2302.14859}}
  (\bibinfo{year}{2023}).
\newblock


\bibitem[Yariv et~al\mbox{.}(2020)]%
        {yariv2020multiview}
\bibfield{author}{\bibinfo{person}{Lior Yariv}, \bibinfo{person}{Yoni Kasten},
  \bibinfo{person}{Dror Moran}, \bibinfo{person}{Meirav Galun},
  \bibinfo{person}{Matan Atzmon}, \bibinfo{person}{Basri Ronen}, {and}
  \bibinfo{person}{Yaron Lipman}.} \bibinfo{year}{2020}\natexlab{}.
\newblock \showarticletitle{Multiview neural surface reconstruction by
  disentangling geometry and appearance}.
\newblock \bibinfo{journal}{\emph{Advances in Neural Information Processing
  Systems}}  \bibinfo{volume}{33} (\bibinfo{year}{2020}),
  \bibinfo{pages}{2492--2502}.
\newblock


\bibitem[Ye et~al\mbox{.}(2023)]%
        {ye2023nef}
\bibfield{author}{\bibinfo{person}{Yunfan Ye}, \bibinfo{person}{Renjiao Yi},
  \bibinfo{person}{Zhirui Gao}, \bibinfo{person}{Chenyang Zhu},
  \bibinfo{person}{Zhiping Cai}, {and} \bibinfo{person}{Kai Xu}.}
  \bibinfo{year}{2023}\natexlab{}.
\newblock \showarticletitle{Nef: Neural edge fields for 3d parametric curve
  reconstruction from multi-view images}. In
  \bibinfo{booktitle}{\emph{Proceedings of the IEEE/CVF Conference on Computer
  Vision and Pattern Recognition}}. \bibinfo{pages}{8486--8495}.
\newblock


\bibitem[Yu et~al\mbox{.}(2022)]%
        {yu2022monosdf}
\bibfield{author}{\bibinfo{person}{Zehao Yu}, \bibinfo{person}{Songyou Peng},
  \bibinfo{person}{Michael Niemeyer}, \bibinfo{person}{Torsten Sattler}, {and}
  \bibinfo{person}{Andreas Geiger}.} \bibinfo{year}{2022}\natexlab{}.
\newblock \showarticletitle{Monosdf: Exploring monocular geometric cues for
  neural implicit surface reconstruction}.
\newblock \bibinfo{journal}{\emph{Advances in neural information processing
  systems}}  \bibinfo{volume}{35} (\bibinfo{year}{2022}),
  \bibinfo{pages}{25018--25032}.
\newblock


\bibitem[Zhang et~al\mbox{.}(2023)]%
        {zhang2023towards}
\bibfield{author}{\bibinfo{person}{Yongqiang Zhang}, \bibinfo{person}{Zhipeng
  Hu}, \bibinfo{person}{Haoqian Wu}, \bibinfo{person}{Minda Zhao},
  \bibinfo{person}{Lincheng Li}, \bibinfo{person}{Zhengxia Zou}, {and}
  \bibinfo{person}{Changjie Fan}.} \bibinfo{year}{2023}\natexlab{}.
\newblock \showarticletitle{Towards unbiased volume rendering of neural
  implicit surfaces with geometry priors}. In
  \bibinfo{booktitle}{\emph{Proceedings of the IEEE/CVF Conference on Computer
  Vision and Pattern Recognition}}. \bibinfo{pages}{4359--4368}.
\newblock


\bibitem[Zhao et~al\mbox{.}(2022)]%
        {zhao2022human}
\bibfield{author}{\bibinfo{person}{Fuqiang Zhao}, \bibinfo{person}{Yuheng
  Jiang}, \bibinfo{person}{Kaixin Yao}, \bibinfo{person}{Jiakai Zhang},
  \bibinfo{person}{Liao Wang}, \bibinfo{person}{Haizhao Dai},
  \bibinfo{person}{Yuhui Zhong}, \bibinfo{person}{Yingliang Zhang},
  \bibinfo{person}{Minye Wu}, \bibinfo{person}{Lan Xu}, {et~al\mbox{.}}}
  \bibinfo{year}{2022}\natexlab{}.
\newblock \showarticletitle{Human performance modeling and rendering via neural
  animated mesh}.
\newblock \bibinfo{journal}{\emph{ACM Transactions on Graphics (TOG)}}
  \bibinfo{volume}{41}, \bibinfo{number}{6} (\bibinfo{year}{2022}),
  \bibinfo{pages}{1--17}.
\newblock


\bibitem[Zhuang et~al\mbox{.}(2023)]%
        {zhuang2023anti}
\bibfield{author}{\bibinfo{person}{Yiyu Zhuang}, \bibinfo{person}{Qi Zhang},
  \bibinfo{person}{Ying Feng}, \bibinfo{person}{Hao Zhu}, \bibinfo{person}{Yao
  Yao}, \bibinfo{person}{Xiaoyu Li}, \bibinfo{person}{Yan-Pei Cao},
  \bibinfo{person}{Ying Shan}, {and} \bibinfo{person}{Xun Cao}.}
  \bibinfo{year}{2023}\natexlab{}.
\newblock \showarticletitle{Anti-aliased neural implicit surfaces with encoding
  level of detail}. In \bibinfo{booktitle}{\emph{SIGGRAPH Asia 2023 Conference
  Papers}}. \bibinfo{pages}{1--10}.
\newblock


\end{thebibliography}

\end{document}


\title{Supplementary Materials: \\VoxNeuS: Enhancing Voxel-Based Neural Surface Reconstruction via Gradient Interpolation}


\author{Anonymous Authors}

\maketitle


\section{Datasets and Baselines}

\section{Additional Experiments}


\subsection{Efficiency of Interpolated Gradients}

\subsection{VoxNeuS with Dual Color Network}

\subsection{Visualizations on BlendedMVS}

\begin{table*}[]
\caption{with mask.}
\begin{tabular}{l|cccccccccccccccc}
\toprule
Scan   & 24   & 37   & 40   & 55   & 63   & 65   & 69   & 83   & 97   & 105  & 106  & 110  & 114  & 118  & 122  & mean \\
\midrule
NeRF\cite{mildenhall2021nerf} & 1.83 & 2.39 & 1.79 & 0.66 & 1.79 & 1.44 & 1.50 & 1.20 & 1.96 & 1.27 & 1.44 & 2.61 & 1.04 & 1.13 & 0.99 & 1.54 \\
IDR\cite{yariv2020multiview} & 1.87 & 1.63 & 0.63 & 0.48 & 1.04 & 0.79 & 0.771 & 1.33 & 1.16 & 0.76 & 0.67 & 0.90 & 0.42 & 0.51 & 0.53 & 0.90 \\
NeuS\cite{wang2021neus} & 0.83 & 0.98 & 0.56 & 0.37 & 1.13 & \textbf{0.59} & \textbf{0.60} & 1.45 & \textbf{0.95} & 0.78 & 0.52 & 1.43 & \textbf{0.36} & 0.45 & 0.45 & 0.77 \\
Voxurf\cite{wu2022voxurf} & 0.65 & \textbf{0.74} & 0.39 & \textbf{0.35} & 0.96 & 0.64 & 0.85 & 1.58 & 1.01 & 0.68 & 0.60 & 1.11 & 0.37 & 0.45 & 0.47 & 0.72 \\
NeuS2\cite{wang2023neus2}  & \textbf{0.56} & 0.76 & 0.49 & 0.37 & \textbf{0.92} & 0.71 & 0.76 & 1.22 & 1.08 & \textbf{0.63} & 0.59 & \textbf{0.89} & 0.40 & 0.48 & 0.55 & \textbf{0.70} \\
\midrule
Ours   & 0.60 & 0.77 & \textbf{0.38} & \textbf{0.35} & 1.27 & 0.75 & 0.75 & \textbf{1.19} & 1.18 & 0.68 & \textbf{0.44} & 0.98 & 0.39 & \textbf{0.42} & \textbf{0.44} & \textbf{0.70} \\
\bottomrule
\end{tabular}
\end{table*}


\bibliographystyle{ACM-Reference-Format}
\bibliography{sample-base}








